\documentclass[10pt,twocolumn,letterpaper]{article}

\usepackage{wacv}      

\definecolor{wacvblue}{rgb}{0.21,0.49,0.74}
\usepackage[pagebackref,breaklinks,colorlinks,allcolors=wacvblue]{hyperref}

\usepackage[table]{xcolor} 
\usepackage{graphicx} 
\usepackage{array}    
\usepackage{multirow} 
\usepackage{cuted}    
\usepackage{caption}  
\usepackage{hyphenat}
\usepackage{algorithm} 
\usepackage{algorithmic}
\providecommand{\linenumbers}{} 
\providecommand{\nolinenumbers}{}

\newcolumntype{C}[1]{>{\centering\arraybackslash}m{#1}}

\newcommand{\method}{LiAM\hyp{}SAM\xspace}
\newcommand{\methoddesc}{\emph{Lifecycle\hyp{}Aware Memory}\xspace}
\newcommand{\methodacc}{LiAM\xspace}

\def\wacvPaperID{1211} 
\def\confName{WACV}
\def\confYear{2027}

\title{\method: Lifecycle-Aware Memory for Robust SAM2-Based MOT}

\author{ Grégoire Francisco\thanks{Contracted services for Toyota Motor Europe.} \and Alessandro D'Amico\footnotemark[1] \and Samuele Costantini \and Gianpiero Francesca \and Lorenzo Garattoni\\[0.5em] Toyota Motor Europe, Zaventem, Belgium }

\begin{document}
\maketitle
\begin{abstract}
Segmentation-based multi-object tracking (MOT) with foundation video models such as SAM2 offers strong localization quality, yet remains fragile in crowded, real-world scenes. 
In detector‑prompted SAM2 pipelines, failures typically arise at three stages of the object lifecycle: (i) erroneous or duplicate track initiation, (ii) memory drift during close interactions, and (iii) unreliable re-identification after long occlusions or re‑entry. These errors corrupt object memory and accumulate over time, making long‑horizon tracking unstable.

In this paper, we reframe MOT as a lifecycle memory integrity problem. We present \method, a \methoddesc (\methodacc) framework with targeted mechanisms for each of the three failure modes. At track birth, to prevent faulty or duplicate initiations, we apply contrastive track initiation, which conditions each prompt on existing nearby tracked instances. To preserve memory integrity during strong interactions, we introduce motion‑ and geometry‑grounded memory correction that resolves interaction confusions and suppresses drift. For reliable re‑identification after disappearance, we maintain an adaptive context memory that promotes diverse and trustworthy references as long‑term identity anchors. Finally, similarity aware spatial pruning optionally selects the memory tokens to retain at cross‑attention time, improving efficiency with minimal accuracy loss.

\method represents a modular, detector-agnostic, SAM2-based MOT system that achieves state-of-the-art HOTA and IDF1 on the evaluated benchmarks. In association-challenging environments, our ablations show that \methodacc improves a detector+SAM2 baseline by +10.5 HOTA, +17.4 AssA, and reduces identity switches by 96\%. 

\end{abstract}    
\section{Introduction}
\label{sec:intro}
Multi-object tracking (MOT) is a fundamental computer-vision task that aims to maintain consistent object identities over time, a core requirement for reliable scene understanding in applications such as autonomous driving, robotics, and surveillance. Recent segmentation-driven trackers built on SAM2~\cite{sam2} improve localization quality, but they remain brittle in crowded and interaction-heavy scenes where frequent overlaps, occlusions, disappearances, and re-entries amplify identity errors.

In detector-prompted SAM2 systems, the main failures arise at three stages of the object lifecycle: (1) \emph{track initiation}, where permissive detections can trigger erroneous or duplicate track births; (2) \emph{interaction-time propagation}, where close contacts can corrupt object-specific memory updates, causing drift, mask bleeding, and identity swaps; and (3) \emph{reappearance}, where long occlusions or exit/re-entry events break re-identification when references are obsolete or weakly curated.

Prior work addresses subsets of this problem, but not the full lifecycle of attention-based memory control. Memory-less trackers such as SORT, ByteTrack, and OC-SORT improve association through motion-consistent updates and low-confidence detection recovery~\cite{sort,bytetrack,ocsort}, yet they do not maintain explicit object memory and therefore degrade under long visibility gaps and complex interactions. Attention-based trackers such as TrackFormer and TransTrack introduce persistent track queries~\cite{trackformer,transtrack}, but their memory has no explicit mechanism to handle severe overlap. Segmentation-centered works based on SAM/SAM2/SAM3, including MOT solutions MASA, SAM2MOT and SAM3~\cite{masa,sam2mot,sam3}, provide stronger mask-level representations, while SAM2-Long and SAMURAI improve long-horizon memory handling~\cite{sam2long,samurai}; however, they do not provide a unified autonomous policy to handle memory corruptions that arise and propagate at the different stages of tracks lifecycle.

These observations motivate our view: robust SAM2-based MOT is a memory-integrity problem. In standard SAM2 usage, memory curation is interactive; in MOT, the same decisions must be made autonomously and online under noisy detections and ambiguous interactions. At the same time, memory-attention complexity grows with both the number of tracked objects and the number of stored references, making memory efficiency increasingly important.

\begin{figure*}[t]
    \centering
    \includegraphics[width=0.88\textwidth]{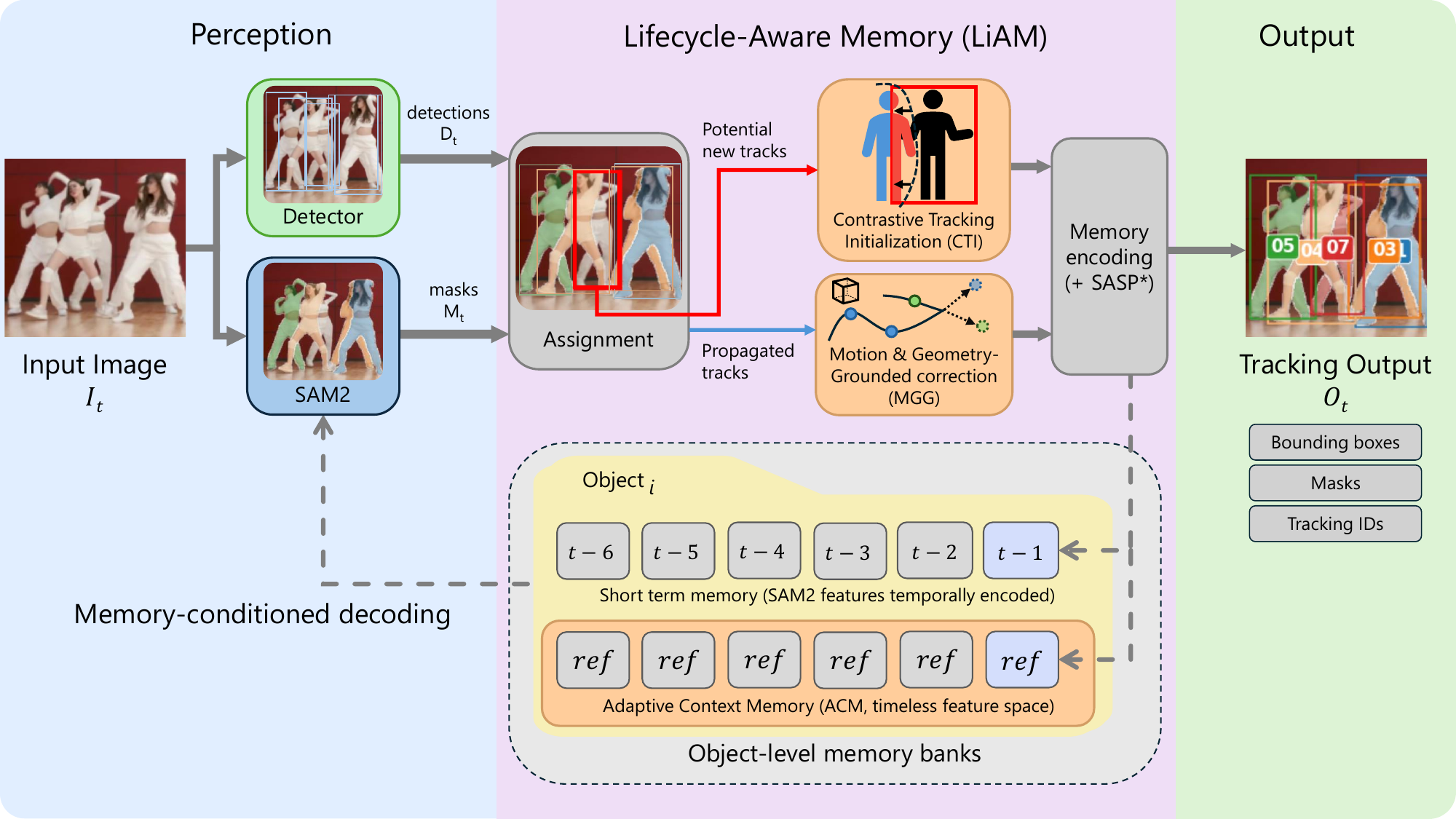}
    \caption{\method{}. Starting from the combination of a detector and SAM2, our method matches bounding boxes and masks and initializes the tracked objects by explicitly presenting to SAM2 negative points---\ie, other tracked objects overlapping with the new potential one  (Contrastive Tracking Initialization). The Adaptive Context Memory (ACM) introduces long memory and is updated only when the appearance is reliable. Finally, Motion \& Geometry Grounded correction corrects confusion during occlusions and mask bleeding using motion and depth cues.}
    \label{fig:method_verview}
\end{figure*}

We address this with \method (\cref{fig:method_verview}), a \methoddesc (\methodacc) framework built on a detector+SAM2 backbone. The central idea is lifecycle-aware memory integrity: regulate what enters memory at track initialization, maintain memory integrity during interactions, and retain reliable references for long\hyp{}horizon re\hyp{}identification. Concretely, \method combines Contrastive Track Initialization (CTI), Motion- and Geometry-Grounded correction (MGG), and Adaptive Context Memory (ACM). We further include Similarity-Aware Spatial Pruning (SASP) as an optional efficiency module that reduces memory-attention cost with limited impact on accuracy.

In summary, our contributions are:
\begin{itemize}
    \item \textbf{A lifecycle-aware memory-integrity formulation for SAM2-based MOT.} We formalize detector-prompted SAM2 tracking as an object-lifecycle memory problem with three main failure stages: track initiation, interaction\hyp{}time propagation, and reappearance.
    \item \textbf{A unified method instantiation in \method.} We instantiate this formulation with CTI for reliable track initiation, MGG for interaction\hyp{}time correction, and ACM for robust long-horizon re-identification.
    \item \textbf{Optional efficiency boost via attention-time pruning.} We include SASP, which prunes spatial memory tokens using shared similarity cues to improve runtime while preserving most tracking quality.
    \item \textbf{Comprehensive empirical validation.} On DanceTrack~\cite{dancetrack}, BDD100K \cite{bdd100k} and SportsMOT~\cite{cui2023sportsmot}, \method delivers state-of-the-art performance and large association gains over a detector+SAM2 baseline, including +10.5 HOTA, +17.4 AssA, and 96\% fewer ID switches on DanceTrack.
\end{itemize}

\section{Related Work}
\label{sec:related}

\subsection{Detection-Based Online Tracking}
Online multi-object tracking (MOT) methods traditionally follow the tracking-by-detection paradigm, where frame-level detections are associated over time to maintain identities. Early methods such as SORT~\cite{sort} rely on Kalman filtering and IoU matching, while DeepSORT~\cite{wojke2017simpleonlinerealtimetracking} adds appearance embeddings to improve association under ambiguity. Later approaches, including JDE~\cite{wang2020realtimemultiobjecttracking} and FairMOT~\cite{fairmot}, jointly learn detection and identity cues, while ByteTrack~\cite{bytetrack} improves robustness by associating both high- and low-confidence detections.
More recent methods such as CenterTrack~\cite{centertrack} and OC-SORT~\cite{ocsort} improve motion continuity: CenterTrack introduces an offset regression branch that predicts displacement vectors between consecutive frames, allowing the tracker to propagate identities without a separate matching stage, while OC-SORT refines Kalman-based updates using observed detection dynamics. Overall, detection-based trackers have evolved from simple motion association toward more integrated and learned propagation schemes.
However, they do not explicitly address the lifecycle of segmentation-based object memory, which becomes critical under long occlusions, re-entry, and close interactions.

\subsection{Attention-Based Trackers}
Attention mechanisms have become prominent in modern MOT systems, enabling robust association across frames by selectively focusing on relevant spatial and temporal cues. These approaches leverage attention to unify detection and tracking within transformer-based architectures.

Parallel approaches (e.g., VisTR~\cite{vistr}, SeqFormer~\cite{seqformer}) apply spatiotemporal attention over short video clips in a single pass, enabling joint reasoning across frames. However, this dense attention incurs high memory costs, restricting scalability to longer sequences.

Serial approaches (e.g., TrackFormer~\cite{trackformer}, TransTrack~\cite{transtrack}) update track queries iteratively, frame by frame. For instance, TransTrack combines attention with IoU-based matching for new detections. 

Recent methods further improve the temporal representation of track queries. ColTrack~\cite{coltrack} retains multiple historical queries for each target to collaboratively refine the current representation, improving robustness to large appearance and displacement changes in low-frame-rate videos. Dual-Path Temporal Decoder~\cite{dualpathtempdec} instead separates appearance adaptation from identity preservation through complementary decoder paths and suppresses unreliable query updates based on prediction confidence.
Despite advances in attention-based approaches, the performance is constrained by limited annotated data and reliance on short-range temporal cues. 

\subsection{Large Foundation Models and SAM2}

SAM2~\cite{sam2} provides strong generalizable video segmentation, making it an attractive backbone for segmentation-driven MOT. Its memory-based architecture stores historical mask-level features encoding appearance and position, supporting identity preservation under occlusions and appearance changes.
Extensions such as SAM2-Long \cite{sam2long} and SAMURAI \cite{samurai} further improve scalability and memory efficiency through hierarchical structures and cleaning strategies.
SAM2MOT~\cite{sam2mot} combines open-vocabulary detections with SAM2 segmentation, but relies on offline video-specific hyperparameter optimization, limiting its practicality for online real-world applications.
SAM3~\cite{sam3} extends the open-vocabulary MOT (OV-MOT) approach to focus on a more efficient and robust open-vocabulary prompter, while SAM2 remains a strong and flexible baseline to adapt any type of detector for more traditional and critical MOT task with a closed set of categories targets.

Building on SAM2, we propose \method, a MOT framework that introduces a robust memory-management system to maintain consistent object identities across long sequences, addressing the limitations of short-range attention in prior methods.
Furthermore, we incorporate depth- and motion-grounded cues to resolve occlusions effectively, enabling the tracker to distinguish overlapping objects and recover identities even under severe visual ambiguity.

\section{Method}
\label{sec:method}
\subsection{Overview}
\label{sec:method:over}
An overview of \method is given in~\cref{fig:method_verview}. We consider monocular multi-object tracking with segmentation. At each frame $t$ with image $I_t$, the tracker $f_\theta$ outputs a set of instances, each represented by a segmentation mask, a bounding box, and a persistent identity.

\method builds on a detector~+~SAM2 tracking loop governed by a \methoddesc (\methodacc). The central idea is \emph{lifecycle-aware memory integrity}: regulate what enters memory at track initiation, protect memory during close interactions, and retain reliable references for long-horizon re-identification. We instantiate this framework with four modules: 
\begin{enumerate*}[label=(\roman*)]
    \item \textbf{Contrastive Track Initialization (CTI)}, which initializes new tracks from detector boxes while suppressing duplicate, false, and sub-object initiations;
    \item \textbf{Adaptive Context Memory (ACM)}, a curated bank of reliable, diverse reference observations for robust long\hyp{}horizon re\hyp{}identification after disappearances and context evolution;
    \item \textbf{Motion- and Geometry-Grounded correction (MGG)}, which resolves identity ambiguities via motion-grounded occlusion handling and suppresses mask drifting via depth-guided bleeding correction; and
    \item \textbf{Similarity-Aware Spatial Pruning (SASP)}, optional token trimming on SAM2 object memory based on shared feature similarity, which reduces cross-attention cost with minimal accuracy impact.
\end{enumerate*}

At each frame $t$, propagated track boxes and detector boxes are associated with the Hungarian algorithm minimizing the IoU cost
\[
\text{cost}_{ik}^t = 1 - \mathrm{IoU}(\tilde{b}_i^t,\, b_k^t),
\]
yielding matched pairs and unmatched detections. The four \methodacc modules act at different lifecycle stages: track birth (CTI), long-term reference curation (ACM), interaction handling (MGG), and attention-time trimming (SASP).



\subsection{Contrastive Track Initialization (CTI)}
\label{sec:method:cti}

MOT is especially sensitive to track birth errors: a false, duplicate, or sub-object initiation can propagate for many subsequent frames. Raising the detector confidence threshold improves birth precision but reduces recall. Instead, CTI keeps detector recall high and enforces a \emph{contrastive admissibility policy} before a new track is accepted.

Let $\{M_i^t\}_{i=1}^{N_t}$ be the masks of active tracks at frame $t$, and let $B$ be an unmatched detector box. Like standard MOT systems, proposals are first pre-filtered for confidence and an overlap-consistency criterion that rejects boxes with excessive occlusion. CTI then uses $B$ as a positive box prompt and, when the candidate overlaps existing tracked content, adds negative point prompts derived from local:
\begin{equation}
    S_i^t(B) = M_i^t \cap B.
\end{equation}
For each non-empty $S_i^t(B)$, we extract a representative point (center of the largest connected component) and form a negative prompt set $\mathcal{P}_B^-$, informing SAM2 that portions of the candidate box are already explained by tracked objects. CTI predicts a candidate mask:
\[
\tilde{M}_B^t = f_\theta(I_t;\, B,\, \mathcal{P}^-_B).
\]
The mask is encouraged to cover the novel object inside $B$ while avoiding regions already claimed by current tracks.

CTI then accepts the new track only if the candidate passes post-initialization consistency checks. The key statistic used for duplicate suppression is the proportion of the proposal mask's area that overlaps existing objects:
\[
\delta_B^t = \frac{|\tilde{M}_B^t \cap M_\cup^t|}{|\tilde{M}_B^t|},
\qquad M_\cup^t = \bigcup_i M_i^t,
\]
which rejects masks largely explained by already-tracked regions. Together, these mechanisms disentangle detection recall from birth precision: permissive detection thresholds are preserved, while spurious births are suppressed through segmentation‑aware contrastive prompting and mask‑level consistency constraints.

\subsection{Adaptive Context Memory (ACM)}
\label{sec:method:acm}

SAM2 predicts masks in subsequent frames using an attention-based decoder conditioned on object memory embeddings. Its short-term memory provides strong temporal continuity, but may become unreliable after long disappearances or context changes (e.g., viewpoint, illumination, or surrounding objects). For each track $i$, we use the standard SAM2 short-term memory of temporally encoded tokens from the last $K$ frames, without modifications.

ACM complements this with a time-unbounded memory of reliable and diverse reference observations, while remaining compatible with SAM2’s native memory.


\textbf{Adaptive context references.}
SAM2 also retains the first memory timestep without temporal encoding as a static object reference.
ACM replaces this original static reference of SAM2 with a curated bank of reliable, evolving past observations stored without temporal encoding. Matched track observations are promoted to this conditioning-memory set when considered reliable, following an update schedule that encourages diversity across evolving contexts. This leads to a compact set of adaptive references that improves robustness to long disappearances and context changes, while preserving the fine-grained continuity of SAM2 short-term memory.

At decoding time, the object query jointly attends to: (i) recent SAM2 short-term memory and (ii) the ACM references. The latter acts as a context-aware re-identification support, especially when the most recent short-term memories are missing or corrupted by interactions.


While other works like  DAM4SAM~\cite{dam4sam} and SAM2‑Long~\cite{sam2long} rely on alternative SAM2‑generated mask proposals to augment their memory banks, our ACM focuses on capturing contextual and modality‑evolving cues, which also broadens its applicability to segmentation backbones that produce only a single mask.

\subsection{Motion- and Geometry-Grounded correction (MGG)}
\label{sec:method:mgg}

Even with CTI and ACM, the most critical failure modes arise during strong interactions between similar objects: duplicate hypotheses can co-exist, masks can bleed across boundaries, and corrupted memories propagate identity switches. MGG addresses these cases by grounding interaction handling in motion and depth cues.

MGG acts on pairs of tracks with spatial overlap using two complementary modules: motion‑grounded occlusion resolution (identity-level ambiguity) and depth‑guided bleeding correction (pixel-level contamination). 

\subsubsection{Motion-Grounded Occlusion Resolution}
\label{sec:method:mgg-motion}

When overlaps are strong, the primary question is often not pixel ownership but \emph{identity validity}: are we observing two distinct colliding objects, or a duplicate hypothesis attached to the same entity? MGG resolves this ambiguity with a motion-grounded occlusion resolver.

For each strongly overlapping pair $(i,j)$, we evaluate short-horizon trajectory distinctness from Kalman-filter motion states, using Mahalanobis-normalized separation over a temporal window. This test distinguishes trajectories that converge into an interaction from nearly indistinguishable duplicate behavior.
If the trajectories are \emph{non-distinct}, MGG treats the pair as duplicates and suppresses the younger track as the less reliable hypothesis. Otherwise, MGG identifies the temporarily unreliable track via score-history degradation, measured as the drop of the current score relative to a recent local baseline. The track with the strongest degradation is selected; if score evidence is inconclusive, MGG falls back to an age-based preference.
The unreliable track persists, but its ambiguous appearance in the current frame is not written to memory: it is excluded from both SAM2 short-term memory and ACM references.
This selective suppression prevents interaction-induced corruption from propagating through memory. If a track is flagged as a confuser in two consecutive frames with the same object, it is considered unrecoverable and aborted to avoid duplicate propagation.
Together, trajectory distinctness, age, and score decay enable stable resolution, whereas scores alone are unreliable.
This motion grounding bears some similarity to SAMURAI’s motion\hyp{}affinity mechanism~\cite{samurai}. However, while SAMURAI generates a motion affinity score to accept/reject proposal at individual\hyp{}object levels, our motion\hyp{}grounding logic preserves memory integrity by distinguishing duplicates and collisions for multiple objects.

\subsubsection{Depth-Guided Bleeding Correction}
\label{sec:method:mgg-depth}

Depth-guided bleeding correction targets moderate overlaps. For a pair $(i,j)$, we define the overlap region $R_{ij}^t = M_i^t \cap M_j^t$ and decompose it into connected components $\{R_{ij,c}^t\}_{c=1}^{N_{ij,c}^{t}}$.
We then estimate the frame depth map $Z_t$ using RGB-monocular-depth-estimation (see~\ref{sex:exp:implementation}).

For each object $k \in \{i,j\}$, we define a support region $S_k^t = M_k^t \setminus R_{ij}^t$ and compute depth statistics on an eroded version of $S_k^t$ to suppress noisy mask edges. Each support region $S_k^t$ yields a support depth interval $I_k^t$ and a dispersion estimate from the restriction of the depth map onto that region ${Z_t|_{S_k^t}}$. Each overlap component $R_{ij,c}^t$ yields a depth interval $I_{ij,c}^t$. We measure how much a component's depth is explained by each object's support via an interval-overlap coverage:
\[
\gamma_{k,c}^t = \frac{|I_k^t \cap I_{ij,c}^t|}{|I_{ij,c}^t|}.
\]

This score explains how much of the overlapping component depth interval is included in the object's support interval.
We accept a depth-based reallocation only when the component is sufficiently explained by one support interval and the two supports are well separated relative to their depth dispersion. In that case, the component is removed from the less coherent mask and the corresponding object memory is immediately re-encoded from the corrected mask, preventing bleed-induced corruption from propagating. If depth evidence is weak or ambiguous, MGG leaves the local overlap unchanged.

\subsection{Similarity-Aware Spatial Pruning (SASP)}
\label{sec:method:sasp}

Finally, we improve the efficiency of SAM2 cross-attention by retaining only the memory tokens that are most relevant for object decoding.  
For each object, SASP keeps tokens from spatial locations inside the predicted mask, as well as tokens from locations with high feature similarity to the object region. These similar locations may correspond to potential confusers and are useful for the decoder to distinguish the target object from visually similar regions. All remaining memory tokens are discarded. The detailed implementation of SASP, together with the theoretical analysis of its complexity reduction, is provided in the supplementary material. Overall, this strategy approximately halves the latency while retaining most of the tracking accuracy.

The overall execution flow of LiAM-SAM is summarized in Algorithm~\ref{alg:liam_update} of the Supplementary Material.

\section{Experiments}

\begin{table*}[htbp]
    \centering
    \scriptsize
    \caption{DanceTrack test-set results. Top block shows general SOTA comparison our RF-DETR based LiAM-SAM. Bottom block provides a detector-ablation, comparing only methods using ByteTrack's YOLOX backbone. Green/yellow indicate best/second-best (within each originally reported table).
    $^\ast$SAM2MOT reports an offline optimization with video-specific thresholds.}
    \label{tab:results_dancetrack_merged}
    \begin{tabular}{@{}l||l| c | c | c | c | c@{}}
    \toprule
        \textbf{Eval} & \textbf{Method} & \textbf{HOTA $\uparrow$} & \textbf{MOTA $\uparrow$} & \textbf{IDF1 $\uparrow$} & \textbf{AssA $\uparrow$} & \textbf{DetA $\uparrow$} \\
    \midrule
        General & FairMOT~\cite{fairmot}            & 39.7 & 82.2 & 40.8 & 23.8 & 66.7 \\
                & CenterTrack~\cite{centertrack}    & 41.8 & 86.8 & 35.7 & 22.6 & 78.1 \\
                & TraDeS~\cite{trades}              & 43.3 & 86.2 & 41.2 & 25.4 & 74.5 \\
                & TransTrack~\cite{transtrack}      & 45.5 & 88.4 & 45.2 & 27.5 & 75.9 \\
                & GTR~\cite{gtr}                    & 48.0 & 84.7 & 50.3 & 31.9 & 72.5 \\
                & QDTrack~\cite{qdtrack}            & 54.2 & 87.7 & 50.4 & 36.8 & 80.1 \\
                & MOTR~\cite{motr}                  & 54.2 & 79.7 & 51.5 & 40.2 & 73.5 \\
                & MeMOTR~\cite{memotr}              & 63.4 & 85.4 & 65.5 & 52.3 & 77.0 \\
                & CO-MOT~\cite{comot}               & 65.3 & 89.3 & 66.5 & 53.5 & 80.1 \\
                & AED~\cite{aed}                    & 66.6 & 92.2 & 69.7 & 54.3 & 82.0 \\
                & MOTIP~\cite{motip}                & 69.6 & 90.6 & 74.7 & 60.4 & 80.4 \\
                & SAMURAI++~\cite{samurai_plus} & 72.8 & 92.1 & 77.7 & 64.7 & 82.2 \\
                & DualTemporalMOT~\cite{dualpathtempdec} & 74.1 & \cellcolor{yellow!20}92.5 & 78.6 & 65.6 & \cellcolor{yellow!20}83.9 \\
                & ColTrack~\cite{coltrack}          & 72.6 & 92.1 & 74.0 & 62.3 & -- \\
                & SAM2MOT~\cite{sam2mot}$^\ast$     & \cellcolor{yellow!20}75.8 & 88.5 & \cellcolor{green!20}83.9 & \cellcolor{yellow!20}72.2 & 79.7 \\
                & SAM2 + RF-DETR & 64.3 & 71.4 & 67.0 & 52.1 & 79.5 \\
                &  \textbf{\method{}}          & \cellcolor{green!20}{\textbf{79.6}} & \cellcolor{green!20}{\textbf{94.3}} & \cellcolor{yellow!20}\textbf{83.6} & \cellcolor{green!20}\textbf{72.6} & \cellcolor{green!20}{\textbf{87.3}} \\
    \midrule\midrule
        Detector  & ByteTrack~\cite{bytetrack}        & 47.7 & 89.6 & 53.9 & 32.1 & 71.0 \\
              Ablation         & OC-SORT~\cite{ocsort}             & 55.1 & 92.0 & 54.6 & 38.3 & 80.3 \\
               (YOLOX)       & StrongSORT++~\cite{strongsort}    & 55.6 & 91.1 & 55.2 & 38.6 & 80.7 \\
                      & C-BIoU~\cite{cbiou}               & 60.6 & 91.6 & 61.6 & 45.4 & 81.3 \\
                      & Hybrid-SORT~\cite{hybridsort}     & 62.2 & 91.6 & 63.0 & 47.4 & 81.9 \\
                      & DualTemporalMOT~\cite{dualpathtempdec} & 67.9 & 87.2 & \cellcolor{yellow!20}73.3 & \cellcolor{yellow!20}60.0 & 77.2 \\
                      & DiffMOT~\cite{diffmot}            & 62.3 & \cellcolor{green!20}92.8 & 63.0 & 47.2 & \cellcolor{yellow!20}82.5 \\
                      & MOTRv2~\cite{motr-v2}             & \cellcolor{yellow!20}69.9 & 92.1 & 71.7 & 59.0 & \cellcolor{green!20}83.0 \\
                      & \shortstack{\textbf{\method{} }\textbf{}}
                        & \cellcolor{green!20}{\shortstack{\textbf{76.5}\textbf{\tiny{(+6.6)}}}}
                        & \cellcolor{yellow!20}{\shortstack{\textbf{92.2}\textbf{\tiny{(-0.6)}}}}
                        & \cellcolor{green!20}{\shortstack{\textbf{83.3}\textbf{\tiny{(+10)}}}}
                        & \cellcolor{green!20}{\shortstack{\textbf{70.6}\textbf{\tiny{(+10.6)}}}}
                        & \shortstack{\textbf{82.0}\textbf{\tiny{(-1.0)}}} \\
    \bottomrule
    \end{tabular}
\end{table*}

\subsection{Datasets and Metrics}
\noindent \textbf{Datasets.} We evaluate \method on three challenging benchmarks with complementary failure modes. \textbf{DanceTrack}~\cite{dancetrack} contains 100 multi-person dance videos (65 train/val, 35 test) with highly similar appearance, non-linear motion, and frequent heavy occlusions, making association quality the dominant challenge. \textbf{BDD100K-MOT}~\cite{bdd100k} targets autonomous driving and contains 1,600 videos (1,400 train, 200 val) across weather, lighting, and scene variations, with annotations for eight traffic-relevant categories (pedestrians, riders, cars, trucks, buses, trains, motorcycles, bicycles). \textbf{SportsMOT}~\cite{cui2023sportsmot} targets multi-object tracking in sports scenarios and contains 240 videos (45 train, 45 val, 150 test) with fast motion, frequent occlusions, and similar appearances, making association particularly challenging.

\noindent \textbf{Metrics.} Our primary metric is HOTA, which balances detection accuracy (DetA) and association accuracy (AssA). We also report MOTA and IDF1 to match common prior reporting for BDD100K-MOT.

\subsection{Implementation details}\label{sex:exp:implementation}
We use RF-DETR-L~\cite{rf_detr}, initialised with COCO-pretrained weights~\cite{coco}, and fine-tune it on the training set of each dataset.
LiAM-SAM itself requires no training: all tracker modules operate with fixed rules and pretrained SAM2.1-large weights, without fine-tuning SAM2 or learning any tracker component. Thus, while the complete detection-and-tracking pipeline uses a dataset-specific detector, the proposed tracker is training-free.
%
Images are resized to a square resolution of $1120 \times 1120$ for training and inference.
We use UniK3D~\cite{unik3d} for monocular depth prediction in MGG-Depth (\cref{sec:method:mgg-depth}). Depth is inferred at the original frame resolution and down-sampled to $256\times 256$. Support/component statistics are computed on eroded masks, which improves robustness to boundary noise.

Detector fine-tuning uses AdamW with effective learning rate $\text{LR}_{\text{eff}} = 1.25 \times 10^{-5}$, encoder learning rate $\text{LR}_{\text{eff,enc}} = 2 \times 10^{-6}$, and weight decay $\text{WD} = 5 \times 10^{-2}$. We train for 10 epochs with physical batch size 4 on DanceTrack and 2 on BDD100K-MOT, with gradient accumulation to an effective batch of 32 samples. We report results on SportsMOT using the same detector and tracking setup.

Data augmentation includes normalization, random horizontal flip ($p=0.5$), multi-scale resizing (long side capped at 1333 pixels, with short-side alternatives 400/500/600), random crops of 384--600 pixels, and final resizing to $1120$. For BDD100K-MOT, we follow the official challenge protocol~\cite{bdd100k_mot_challenge2020} and fine-tune the detector using only the MOT train split. For more details see~\cref{supp:detector}.

\subsection{Comparison with state-of-the-art methods}

\noindent \textbf{DanceTrack (test set).} Table~\ref{tab:results_dancetrack_merged} reports the main comparison on DanceTrack test. \method achieves the best HOTA (79.55), MOTA (94.33), AssA (72.59), and DetA (87.25) among all listed methods, while remaining competitive in IDF1 (83.6, second only to the offline SAM2MOT with a 0.3 point difference). Relative to the naive SAM2+RF-DETR baseline, \method improves HOTA by +15.25, AssA by +20.49, and reduces identity switches by 96\%, showing that the gain is primarily driven by stronger association rather than proposal quality alone.

\noindent \textbf{Detector-controlled comparison (YOLOX-reported methods).}
In Table~\ref{tab:results_dancetrack_merged}, lower block, we evaluate \method using a YOLOX detector and compare against prior YOLOX-based methods. Specifically, we use the official fine-tuned YOLOX checkpoint released by ByteTrack, without any additional fine-tuning. Using this detector, \method achieves the best HOTA (76.5), IDF1 (83.3), and AssA (70.6). In particular, it improves over the previous best HOTA by +6.6 points, while increasing IDF1 and AssA by +10.0 and +10.6 points, respectively, indicating that the gains primarily originate from \methodacc.

\begin{table}[!htbp]
\centering
\scriptsize
    \caption{Comparison on BDD100K validation set.}
    \label{tab:results_bdd}
\begin{tabular}{@{}l| c  | c@{}}
\toprule
Method   & \textbf{IDF1$\uparrow$}& \textbf{MOTA $\uparrow$} \\ \toprule
Unicorn \cite{unicorn} & 71.3 & 66.6\\
QDTrack \cite{qdtrack} & 71.5 & 63.5 \\
SAM2MOT \cite{sam2mot} & 70.8 & 57.5 \\
MOTR-v2 \cite{motr-v2} & 72.7 & 65.6 \\
UNINEXT \cite{uninext}  & 69.9 & \textbf{67.1} \\
\midrule
\method (ours)  & \textbf{73.4} & 64.8 \\
\bottomrule
\end{tabular}
\end{table}

\noindent \textbf{BDD100K\hyp{}MOT (validation set).} Table~\ref{tab:results_bdd} shows the comparison on BDD100K\hyp{}MOT validation. \method achieves the best IDF1 (73.4), improving over MOTR-v2 (72.7) while using a training-free tracker on top of detector inputs. Our MOTA (64.8) is below UNINEXT (67.1), which is expected since UNINEXT prioritizes stronger detection performance with a substantially larger backbone.

\begin{table}[t]
\centering
\setlength{\tabcolsep}{4pt}
\renewcommand{\arraystretch}{0.7}
\scriptsize
\caption{Comparison on SportsMOT test set.}
\label{tab:sportsmot_test}
\begin{tabular}{@{}l| c | c | c | c@{}}
\toprule
Method  & \textbf{HOTA$\uparrow$}  & \textbf{IDF1$\uparrow$}  & \textbf{AssA$\uparrow$}  & \textbf{DetA$\uparrow$} \\ \toprule
ByteTrack      & 62.8 & 69.8 & 51.4 & 77.1 \\
MixSort-OC     & 74.1 & 74.4 & 62.0 & 88.5 \\
Deep-EIoU       & 77.2 & 79.8 & 67.7 & 88.2 \\
DualTemporalMOT & 73.9 & 78.7 & 66.6 & 82.2 \\
Deep HM-SORT  & 80.1 & 85.2 & 72.7 & \textbf{88.3} \\
\midrule
\textbf{LiAM-SAM (ours)} & \textbf{81.8} & \textbf{86.4} & \textbf{76.7} & 87.3 \\
\bottomrule
\end{tabular}
\end{table}

\noindent \textbf{SportsMOT (test set).} 
To further evaluate generalization to challenging motion dynamics, we evaluate LiAM-SAM on the SportsMOT (Tab.~\ref{tab:sportsmot_test}). LiAM-SAM achieves the best HOTA (81.8), IDF1 (86.4), and AssA (76.7), outperforming the previous state-of-the-art Deep HM-SORT. In particular, the +4.0 gain in AssA highlights substantial improvements in association quality, which is consistent with what was observed on DanceTrack and BDD100K.
%
Qualitative examples are provided in Section~\ref{sec:sup:quality}.

\nolinenumbers
\begin{figure*}[t]
\vspace{-1em}
\centering

\setlength{\tabcolsep}{2pt}
\renewcommand{\arraystretch}{0.9}
\newcommand{\qimg}[1]{\makebox[\linewidth][c]{\includegraphics[width=0.97\linewidth]{#1}}}
\begin{tabular}{
    @{}C{0.03\textwidth}@{}  
    C{0.02\textwidth}@{}     
    C{0.27\textwidth}@{}      
    C{0.27\textwidth}@{}      
    C{0.27\textwidth}@{}      
}

\multirow{2}{*}{\rotatebox[origin=c]{90}{BDD100K}}%
& \rotatebox[origin=c]{90}{Ours}%
  & \makebox[\linewidth][c]{\includegraphics[width=0.80\linewidth]{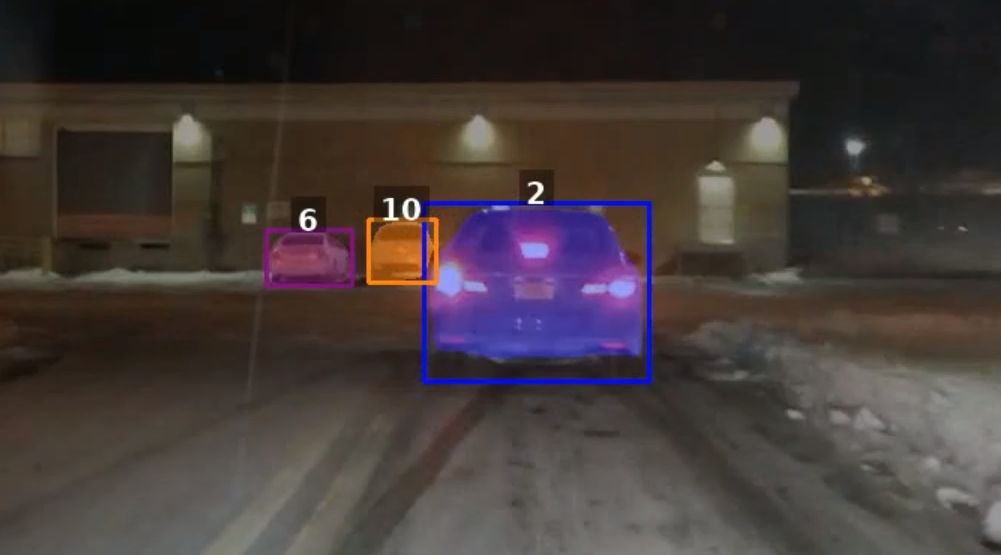}}%
  & \makebox[\linewidth][c]{\hspace{-3pt}\includegraphics[width=0.80\linewidth]{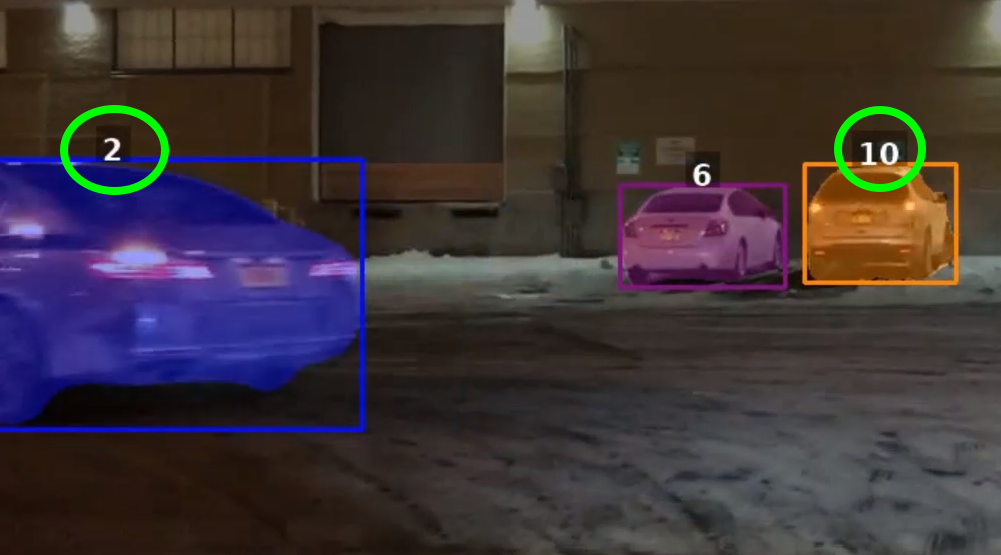}}%
  & \makebox[\linewidth][c]{\includegraphics[width=0.80\linewidth]{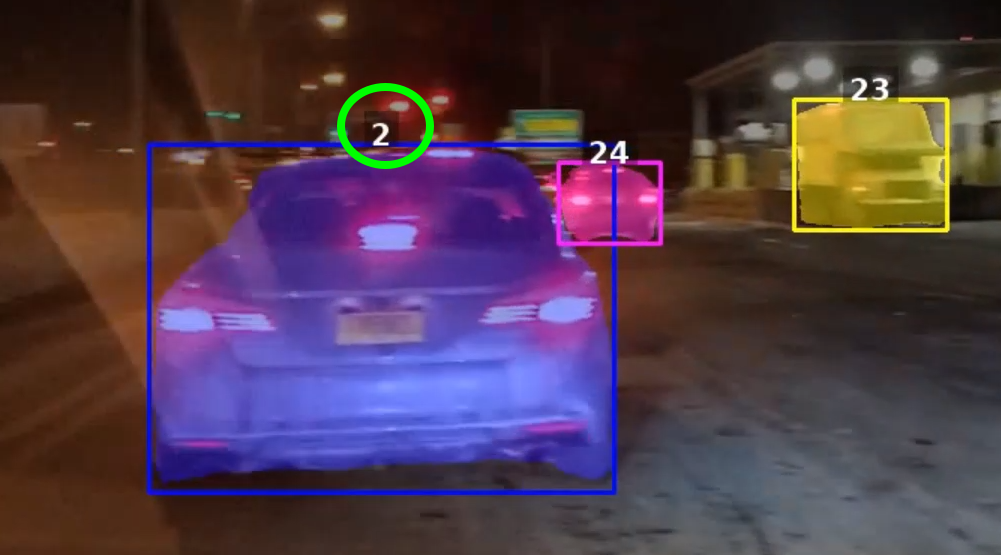}}\\[0.0em]

& \rotatebox[origin=c]{90}{Baseline}%
 & \parbox[c]{\linewidth}{%
    \centering 
    \includegraphics[width=0.80\linewidth]{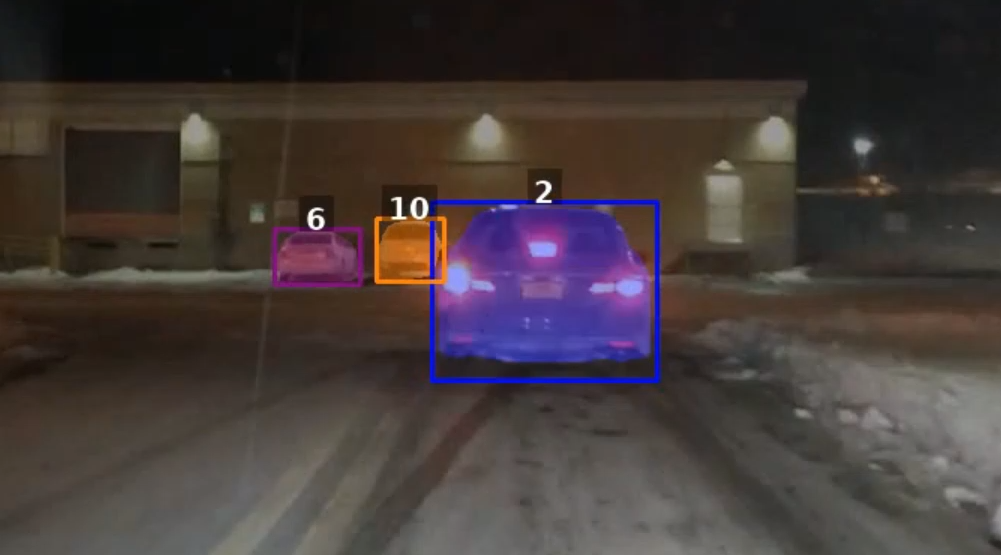}\\
    [-0.2em]
     {\scriptsize (1.a) Pre-disappearance\\[0.47em]}%
     } 
  & \parbox[c]{\linewidth}{%
  \centering 
  \hspace*{-3pt}\includegraphics[width=0.80\linewidth]
{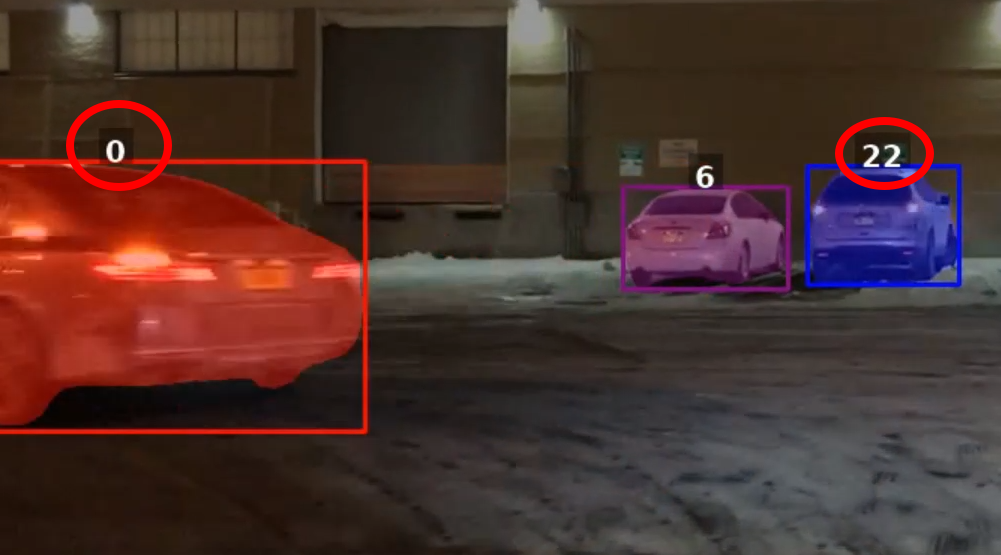}\\[-0.2em] 
  {\scriptsize (1.b) ID loss (\#10$\to$22) and ID swap (\#2$\to$0)\\ 
  caused by complex crossing}%
  } 
  & \parbox[c]{\linewidth}{%
  \centering 
  \includegraphics[width=0.80\linewidth]{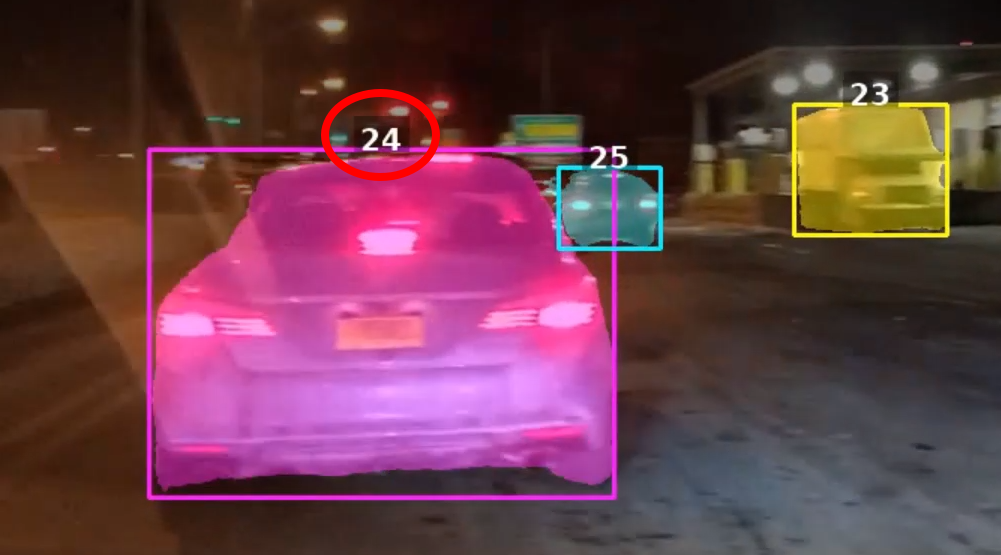}
  \\
  [-0.2em] 
  {\scriptsize (1.c) ACM enables re-ID after reappearance\\[0.47em]}%
}\\
\noalign{\vspace{0.2em}}

\multirow{2}{*}{\rotatebox[origin=c]{90}{DanceTrack}}%
  & \rotatebox[origin=c]{90}{Ours}%
  & \makebox[\linewidth][c]{\includegraphics[width=0.80\linewidth]{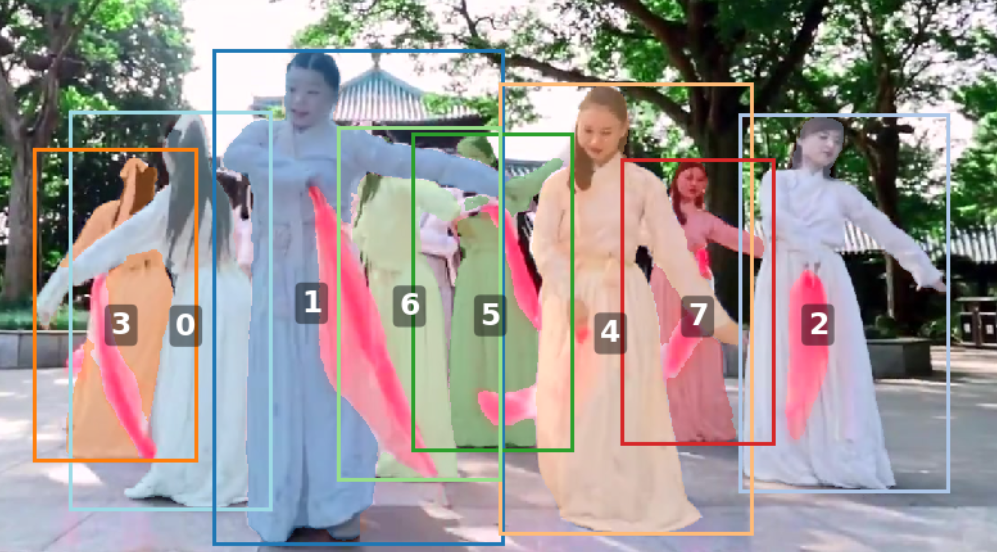}}%
  & \makebox[\linewidth][c]{\hspace{-4pt}\includegraphics[width=0.80\linewidth]{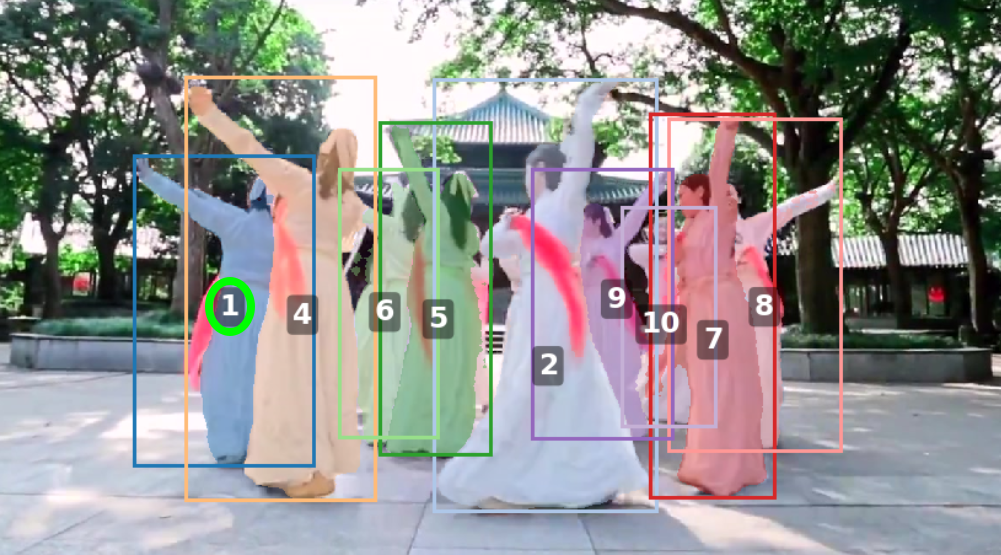}}%
  & \makebox[\linewidth][c]{\includegraphics[width=0.80\linewidth]{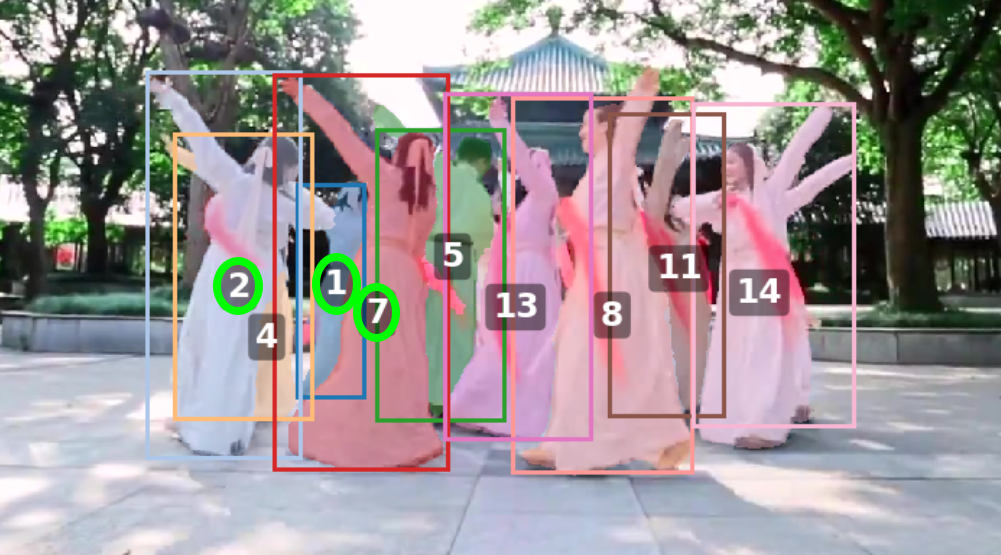}}\\[0.0em]
& \rotatebox[origin=c]{90}{Baseline}%
& \parbox[c]{\linewidth}{%
\centering 
\includegraphics[width=0.80\linewidth]{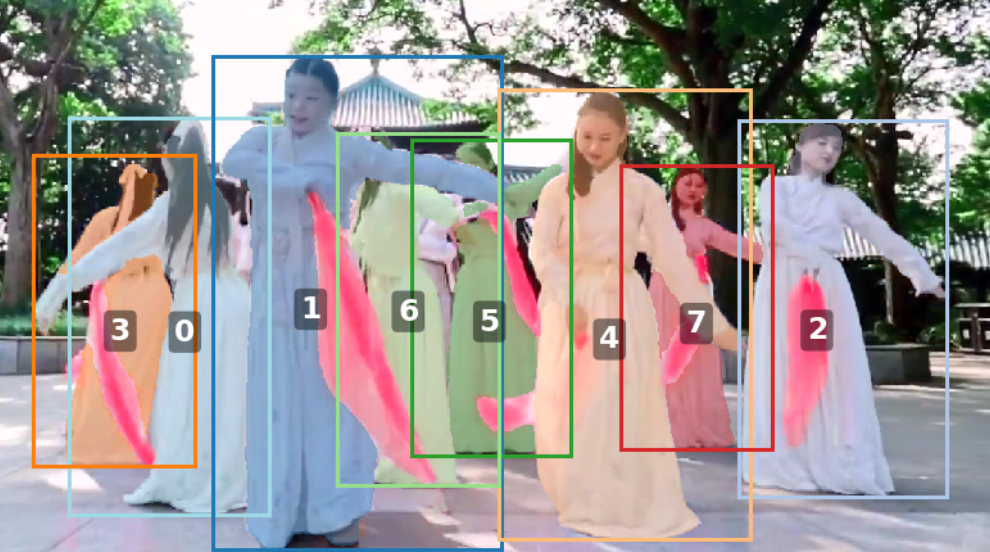}\\
[-0.2em] 
{\scriptsize (2.a) Same initialization\\[0.47em]}%
 } 
& \parbox[c]{\linewidth}{%
\centering 
\hspace*{-4pt}\includegraphics[width=0.80\linewidth]
{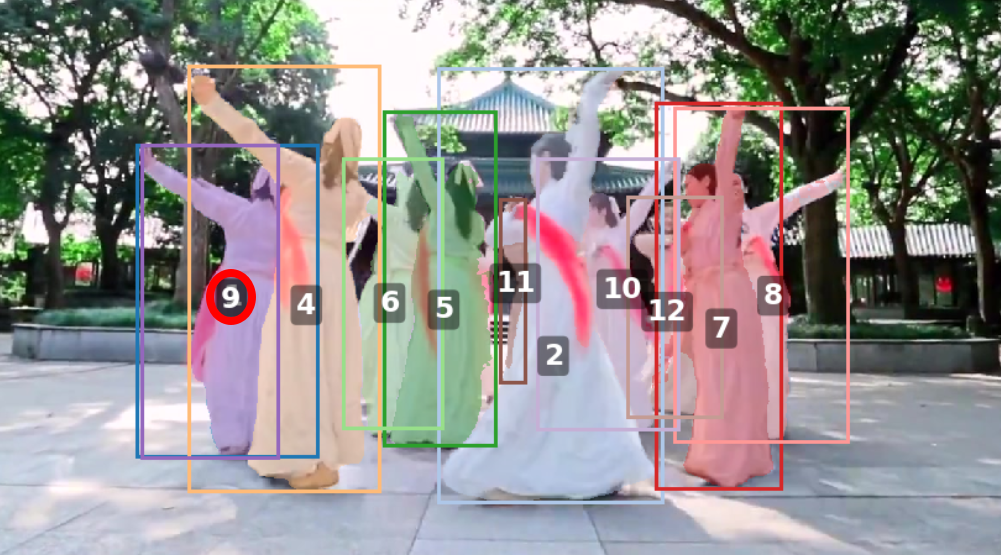}\\
[-0.2em] 
{\scriptsize (2.b) CTI avoids duplicates (\#9$\to$1)\\[0.47em]}%
 } & \parbox[c]{\linewidth}{%
 \centering
 \hspace*{-5pt}\includegraphics[width=0.80\linewidth]
 {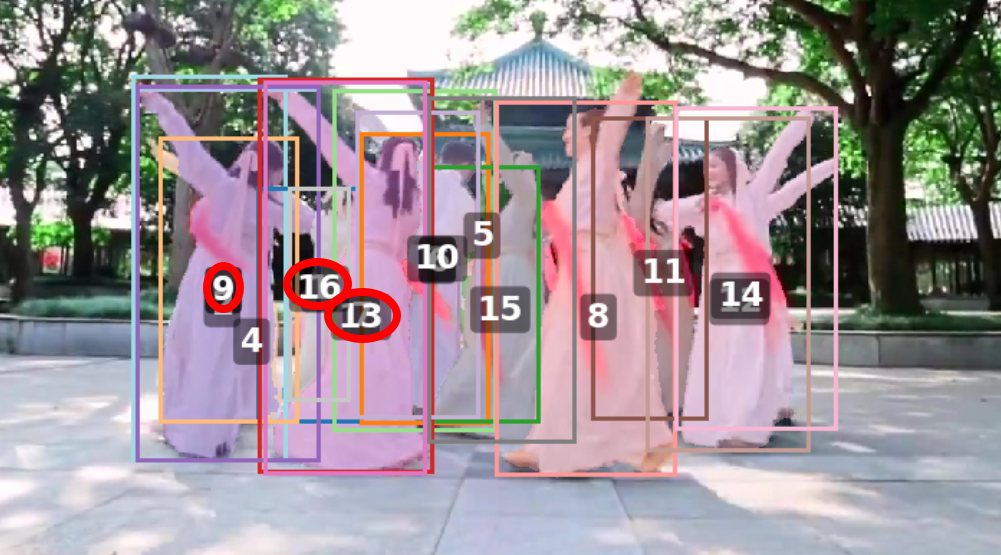}\\[-0.2em] 
 {\scriptsize (2.c) Motion-grounding avoids ID-swaps (\#9$\to$2)\\ CTI filters further duplicates (\#16$\to$1, \#13$\to$7)}%
 }\\
\noalign{\vspace{0.2em}}

\multirow{3}{*}{\rotatebox[origin=c]{90}{SportsMOT}}%
  & \rotatebox[origin=c]{90}{Ours}%
 & \makebox[\linewidth][c]{\includegraphics[width=0.80\linewidth]{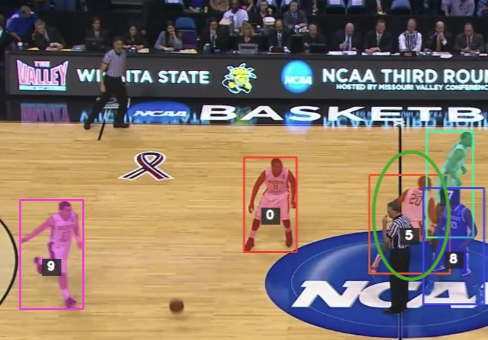}}%
& \makebox[\linewidth][c]{\hspace{-4pt}\includegraphics[width=0.80\linewidth]{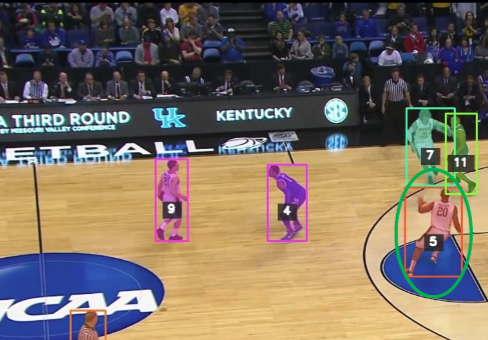}}%
& \makebox[\linewidth][c]{\includegraphics[width=0.80\linewidth]{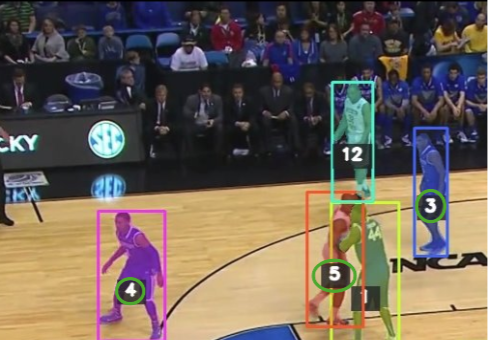}}\\[0.0em]

& \rotatebox[origin=c]{90}{Baseline}%
  & \makebox[\linewidth][c]{\includegraphics[width=0.80\linewidth]{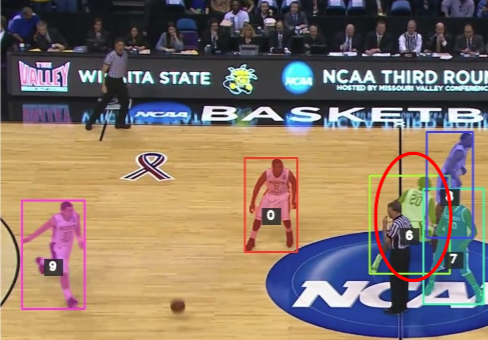}}%
  & \makebox[\linewidth][c]{\hspace{-4pt}\includegraphics[width=0.80\linewidth]{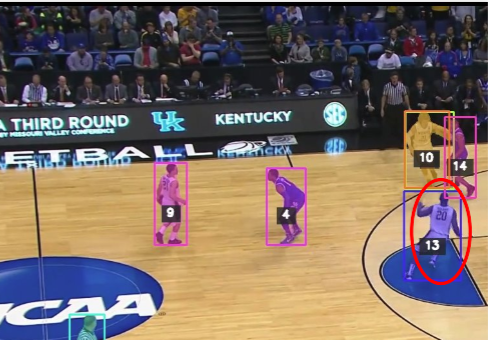}}%
  & \makebox[\linewidth][c]{\hspace{-1pt}\includegraphics[width=0.80\linewidth]{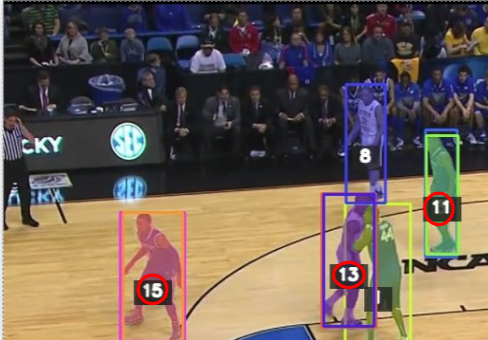}}\\[-0.2em]

& %
  & \multicolumn{2}{c}{%
      \parbox{0.54\textwidth}{%
      \centering\scriptsize
      (3.a--b) Baseline switches ID \#6$\to$\#13 after occlusion; ACM preserves identity.
      }%
    }%
  & \parbox{0.27\textwidth}{%
      \centering\scriptsize
      (3.c) CTI avoid duplicates
    }\\

\end{tabular}

\caption{Qualitative examples of \method on BDD100k, DanceTrack and SportsMOT. See Section \ref{sec:sup:quality}}
\label{fig:quality_dance}
\vspace{-1em}
\end{figure*}
\linenumbers

\subsection{Ablation studies}
Table~\ref{tab:ablation} reports the component ablation on DanceTrack validation. Beyond HOTA\hyp{}IDF1\hyp{}MOTA, we report AssA and IDSW to highlight the association gains of each component.

\begin{table*}[htbp]
\centering
\tiny
\caption{Component ablation on DanceTrack validation set. In addition to HOTA/IDF1/MOTA, we report AssA and IDSW.}
\label{tab:ablation}
\begin{tabular}{l|cccc|c|c|c|c|c|c}
\toprule
\textbf{Method} & \textbf{ACM} & \textbf{CTI} & \textbf{Mo} & \textbf{De} & \textbf{HOTA $\uparrow$} & \textbf{IDF1 $\uparrow$} & \textbf{MOTA $\uparrow$} & \textbf{DetA $\uparrow$} & \textbf{AssA $\uparrow$} & \textbf{IDSW $\downarrow$} \\
\midrule
A: Baseline & -- & -- & -- & -- & 68.0 & 69.6 & 76.3 & 83.9 & 55.3 & 36210 \ \ \ \ \ \ \ \ \ \ \ \ \ \ \ \  . \\
B: A + ACM             & \checkmark & -- & -- & -- & 71.2 & 73.5 & 78.8 & 84.1 & 60.4 (+5.0 vs.A) & 30753 (-15.1\% vs.A) \\
C: A + CTI             & -- & \checkmark & -- & -- & 71.9 & 74.5 & 83.5 & 83.9 & 61.8 (+6.4 vs.A) & 19987 (-44.8\% vs.A) \\
D: C + ACM             & \checkmark & \checkmark & -- & -- & 75.7 & 80.0 & 86.6 & 84.1 & 68.2 (+6.4 vs.C) & 13407 (-32.9\% vs.C) \\
E: B + Mo              & \checkmark & -- & \checkmark & -- & 76.1 & 80.4 & 92.1 & 84.3 & 68.8 (+8.4 vs.B) & \ 1546 (-95.0\% vs.B) \\
F: C + Mo              & -- & \checkmark & \checkmark & -- & 77.7 & 82.5 & 92.1 & 84.9 & 71.1 (+9.3 vs.C) & \ 2435 (-87.8\% vs.C) \\
G: D + Mo              & \checkmark & \checkmark & \checkmark & -- & 78.2 & 83.3 & 92.4 & 84.8 & 72.2 (+4.0 vs.D) & \ 1456 (-89.1\% vs.D) \\
H: G + De              & \checkmark & \checkmark & \checkmark & \checkmark & 78.5 & 83.6 & 92.4 & 84.8 & 72.7 (+0.4 vs.G) & \ 1416 (\ -2.7\% vs.G) \\
\hline
\end{tabular}

\text{Baseline: SAM2+RF-DETR. ACM: Adaptive Context Memory. CTI: Contrastive Tracking Initialization.}
\text{Mo: motion component of MGG. De: depth component of MGG.}
\end{table*}

\begin{table}[h]
\centering
\scriptsize
\caption{
SASP ablation on DanceTrack validation. FPS measured on an NVIDIA A100 GPU. Depth grounding disabled.
}
\label{tab:ablation_pruning}
\begin{tabular}{l|c|c|c}
\toprule
\textbf{Method} & \textbf{HOTA $\uparrow$} & \textbf{FPS-4 obj $\uparrow$} & \textbf{FPS-30 obj$\uparrow$}\\
\midrule
\method        &  78.2 & 5 & 2 \\
LiAM + SASP       & 76.3 & 10 & 5 \\
\hline
\end{tabular}
\tiny
\end{table}

\noindent \textbf{CTI vs. ACM (without DeMo).} Both CTI and ACM improve the naive SAM2+RFDETR baseline (A), but they affect the metrics differently. CTI (C) yields a stronger early gain than ACM-only (B), improving HOTA from 68.0 to 71.9 and reducing IDSW by 44.8\% (vs.\ 15.1\% for ACM-only). This is consistent with CTI's role in reducing duplicate and sub-object track births. ACM-only provides a complementary improvement, especially in AssA (55.3 $\rightarrow$ 60.4), showing that reliable context references already strengthen re-identification.

\noindent \textbf{Combining CTI and ACM.} Their combination (D) is additive: HOTA reaches 75.7, AssA reaches 68.2, and IDSW drops to 13,407 (i.e. a -33\% reduction from 36,210 in the baseline - row A). This confirms that better birth decisions (CTI) and stronger context references (ACM) address different failure modes.

\noindent \textbf{MGG-Motion component (Mo): main association gain.} Adding the motion component of MGG produces the largest reductions in ID switches across all settings. In rows E/F/G, IDSW drops by 95.0\%, 87.8\%, and 89.1\% relative to their corresponding parent configurations (B/C/D), while AssA increases by +8.4, +9.3, and +4.0, respectively. This pattern highlights that overlap-time duplicate/confuser suppression remains a major source of association error.

\noindent \textbf{MGG-Depth component (De): final refinement.} Adding the depth component on top of the full CTI+ACM+Mo configuration (G $\rightarrow$ H) brings a smaller but consistent final gain, improving AssA (72.2 $\rightarrow$ 72.7) and further reducing IDSW by 2.7\%. This matches its intended role as a local mask/memory refinement during moderate overlap, rather than a primary driver of duplicate suppression.
Finally, Table~\ref{tab:ablation_pruning} reports the runtime/accuracy trade-off on DanceTrack validation. Adding \textbf{SASP} to LiAM-SAM increases FPS from 5 to 10 for 4 objects and from 2 to 5 for 30 objects (approximately $2.5\times$ in the crowded case), while preserving 76.3 HOTA, i.e., about $97\%$ of the full-model performances. A full runtime analysis is in Supplementary~\cref{tab:latency_breakdown}.


\subsection{Qualitative Results}
\label{sec:sup:quality}
Figure~\ref{fig:quality_dance} compares \method with a SAM2 + RF-DETR baseline on BDD100K, DanceTrack, and SportsMOT. More examples are given in the Supplementary and video.

\textit{Driving scenario: preserving identity persistence.}
Panel~1 shows long occlusion and re-entry. After car~\#2 is occluded by cars~\#6 and~\#10, the baseline breaks identity (new track~\#22) and later mismatches car~\#2 to track~\#0. \method keeps a single identity through occlusion, viewpoint change, temporary exit, and re-entry, consistent with ACM's long-context references.

\textit{Challenging crowds: reducing duplicates and switches.}
Panel~2 illustrates interaction-heavy dancers with similar appearance. Ambiguous detections create duplicate tracklets in the baseline (\#9, \#13, \#14) and trigger identity swaps (\#9$\rightarrow$\#2 and \#1$\rightarrow$\#8). In contrast, CTI suppresses duplicate births and motion-grounded MGG resolves overlap-time confusers, preserving identity continuity.

\textit{Sports scenario: identity and duplicate control.}
Panel~3 shows a fast-moving sports sequence with occlusion and re-entry. 
Baseline switches ID~\#6 to \#13 after occlusion, whereas LiAM-SAM preserves identity through ACM re-identification support and prevents duplicate births via CTI.

\subsection{Limitations}
Although \method shows state-of-the-art performance across datasets, two practical considerations remain. First, its segmentation‑driven memory operations introduce a computational overhead relative to lighter tracking pipelines. SASP provides an initial step toward mitigating this overhead, though additional optimizations would be required to reach real‑time operation in densely crowded scenes. Second, MGG-depth, which relies on RGB‑derived depth estimates, may be affected in scenes where monocular depth inference is particularly unstable. In practice, these factors had limited impact on our results, and the modular design leaves room for further improvements.

\section{Conclusions}
\label{sec:conclusions}
We presented \method, a segmentation-based MOT framework built on a detector+SAM2 backbone and organized with a \methoddesc (\methodacc) framework. The method combines four complementary modules---CTI, ACM, MGG, and SASP---that act on track birth, reference quality, interaction handling, and memory-attention efficiency, respectively.
Across DanceTrack, BDD100K and SportsMOT, the proposed design yields strong improvements over a SAM2+detector baseline and state-of-the-art performance compared to prior methods. The ablation studies clarify the role of each component to the performance.
Overall, \method demonstrates that coupling segmentation‑based models with principled lifecyle-aware memory leads to substantial robustness gains in multi‑object tracking. We believe the modular nature of the framework also offers a flexible foundation for future extensions, including improved long‑range memory mechanisms and richer multi‑modal cues.


{
    \small
    \bibliographystyle{ieeenat_fullname}
    \bibliography{main}
}

\clearpage

\maketitlesupplementary

\section{SportsMOT Ablation Study} \label{sec:sportsmot_ablation}
Table~\ref{tab:sportsmot_ablation} reports the additional ablation study on SportsMOT validation without tuning individual rows. All rows cover all SportsMOT validation videos and use the submitted RF-DETR checkpoint and tracker
settings. The results are consistent with those reported for DanceTrack in Table~\ref{tab:ablation} of the paper, demonstrating that LiAM’s improvements generalize across datasets and domains.

\begin{table}[ht] \centering \vspace{-10pt}\caption{Ablation study on SportsMOT validation.} \label{tab:sportsmot_ablation} \setlength{\tabcolsep}{1.9pt} \vspace{-10pt} \renewcommand{\arraystretch}{0.80} \fontsize{5.7}{6.2}\selectfont \begin{tabular}{l|cccc|rrr} \toprule & ACM & CTI & Mo & De & HOTA & AssA & IDSW \\ \midrule A: Baseline & $-$ & $-$ & $-$ & $-$ & 73.4 & 61.2 & 36{,}242 \\ B: A + ACM & $\checkmark$ & $-$ & $-$ & $-$ & 78.1 & 69.1 & 32{,}307 \\ C: A + CTI & $-$ & $\checkmark$ & $-$ & $-$ & 76.3 & 66.2 & 20{,}341 \\ D: C + ACM & $\checkmark$ & $\checkmark$ & $-$ & $-$ & 81.9 & 76.2 & 14{,}278 \\ E: B + Mo & $\checkmark$ & $-$ & $\checkmark$ & $-$ & 80.9 & 74.7 & 712 \\ F: C + Mo & $-$ & $\checkmark$ & $\checkmark$ & $-$ & 84.2 & 80.4 & 1{,}128 \\ G: D + Mo & $\checkmark$ & $\checkmark$ & $\checkmark$ & $-$ & 84.3 & 80.7 & 547 \\ H: G + De & $\checkmark$ & $\checkmark$ & $\checkmark$ & $\checkmark$ & 84.3 & 80.8 & 549 \\ \bottomrule \end{tabular}\vspace{-10pt}\end{table}

\section{Similarity-Aware Spatial Pruning (SASP)}
\label{supp:sasp:desc}

In SAM2, each frame $t$ is encoded as image-embedding tokens
$\mathbf{f}_\ell^t \in \mathbb{R}^d$, $\ell = 1,\dots,L$, where
$L = 64 \times 64 = 4096$. For each track $i$, SAM2 fuses
$\mathbf{f}_\ell^t$ with object-specific mask logits via a ConvNeXt-style
block to produce object memory tokens
\begin{equation}
    \mathbf{m}_{i,\ell}^t = g\!\left(\mathbf{f}_\ell^t,\, \alpha_{i,\ell}^t\right) \in \mathbb{R}^d,
\end{equation}
where $\alpha_{i,\ell}^t$ is the mask logit of object $i$ at location $\ell$.
The decoder then cross-attends from object queries to the full memory sequence
$\{\mathbf{m}_{i,\ell}^t\}_{\ell=1}^L$, which is the dominant cost in crowded
scenes.

SASP is the token-pruning stage of \methodacc. For each object memory
$\{\mathbf{m}_{i,\ell}^t\}_\ell$, it keeps:
(i) tokens on the object itself, and
(ii) tokens in regions visually similar to the object (likely confusers).
Similarity is computed in the shared image-embedding space
$\{\mathbf{f}_\ell^t\}$ and reused across all objects.

\paragraph{Frame-level token similarity.}
Once per frame, we compute an absolute cosine-similarity matrix over
image-embedding tokens:
\begin{equation}
    \mathrm{sim}^t(\ell,\ell')
    = \left|
    \frac{\langle \mathbf{f}_\ell^t,\, \mathbf{f}_{\ell'}^t \rangle}
         {\|\mathbf{f}_\ell^t\|_2\,\|\mathbf{f}_{\ell'}^t\|_2}
    \right|, \quad \ell,\ell' \in \{1,\dots,L\}.
\end{equation}
This matrix is shared across all objects, so its overhead is small relative to
the cross-attention savings.

\paragraph{Object seed patches.}
For track $i$ at time $t$, let $\hat{M}_i^t$ denote the (dilated) predicted
mask used for pruning. The \emph{seed patch set} is
\begin{equation}
    \mathcal{P}_i^t = \big\{ \ell \;\big|\; \ell \in \hat{M}_i^t \big\}.
\end{equation}

\paragraph{Similarity-based expansion and pruning.}
We keep all tokens $\mathcal{S}_i^t$ that are sufficiently similar to at least
one seed token:
\begin{equation}
    \mathcal{S}_i^t =
    \Big\{ \ell \;\Big|\; \exists\,\ell_0 \in \mathcal{P}_i^t
    \ \text{s.t.}\ \mathrm{sim}^t(\ell_0,\ell) \ge \tau_\text{sim} \Big\},
\end{equation}
where $\tau_\text{sim}$ is a fixed threshold. Using absolute cosine similarity
lets SASP retrieve both positively and negatively correlated confuser regions.
The pruned object memory used in cross-attention is
\begin{equation}
    \tilde{\mathcal{M}}_i^t = \big\{ \mathbf{m}_{i,\ell}^t \big\}_{\ell \in \mathcal{S}_i^t}.
\end{equation}
Decoupling similarity computation (shared $\mathbf{f}_\ell^t$) from token
selection (object-specific $\mathbf{m}_{i,\ell}^t$) makes SASP agnostic to the
fusion block $g(\cdot)$ and naturally reusable across objects. In practice,
$|\mathcal{S}_i^t| \approx 0.4L$, giving about $60\%$ token reduction per object
per frame, about $2.5\times$ cross-attention speedup, and about $2\times$ FPS
improvement (see \cref{sec:experiments:pruning}), while retaining about
$97\%$ of full-model tracking accuracy.

\section{SASP: From FLOPs to Latency}
\label{sec:experiments:pruning}
This section quantifies SASP from a FLOP perspective. If pruning keeps
approximately $0.4L$ tokens, the ideal cross-attention reduction is
$1/0.4 \approx 2.5\times$. In practice, wall-clock gains can be larger due to
improved memory access and GPU kernel behavior. The similarity map is computed
once per frame and reused across objects, so its overhead remains small.

\begin{figure}[ht]
    \centering
    \includegraphics[width=0.4\textwidth]{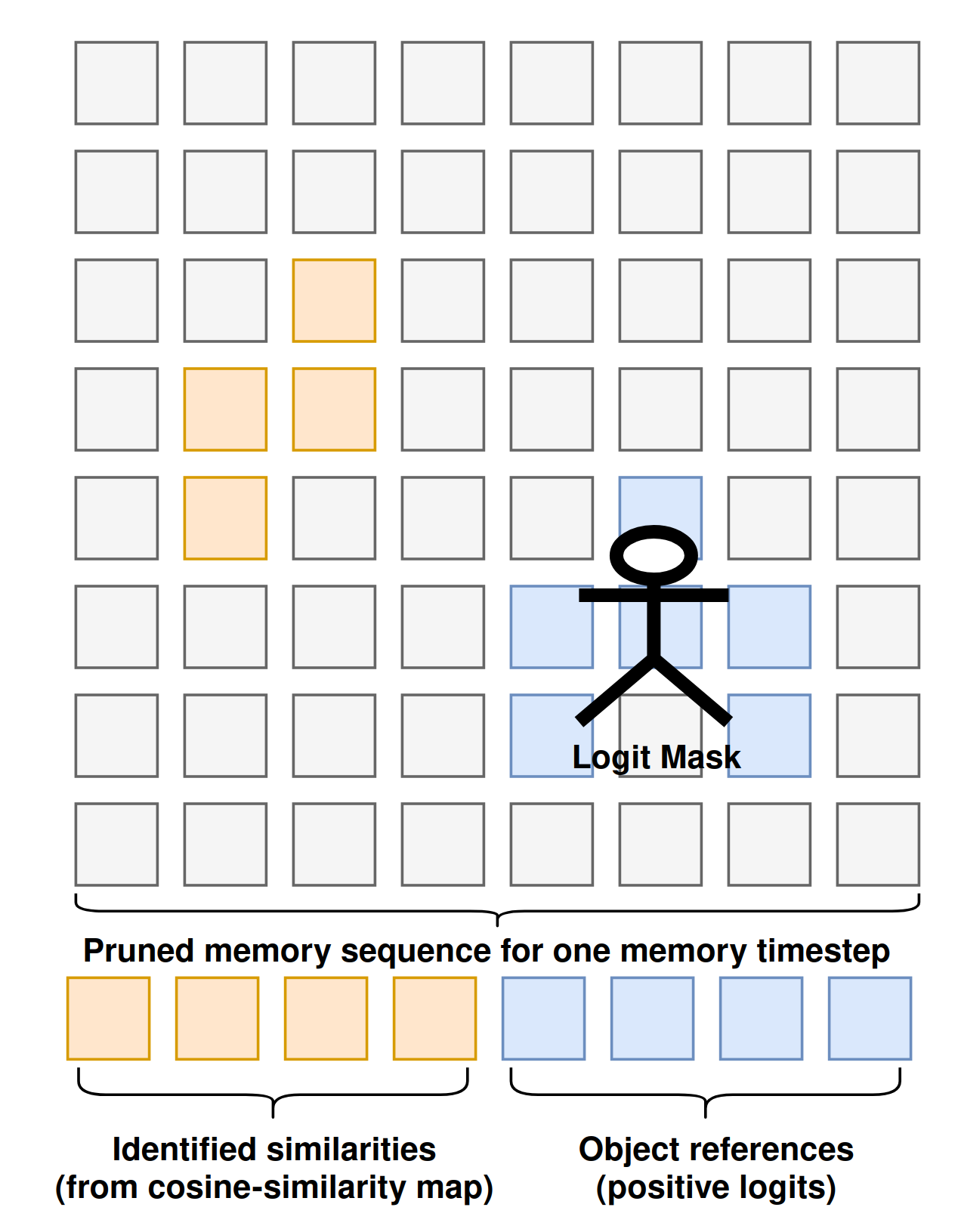}
    \caption{Similarity-Aware Spatial Pruning (SASP). The gray grid denotes all
    memory tokens for one object at one timestep. SASP keeps tokens on the
    object mask (blue) and tokens at visually similar locations in the shared
    image-embedding space (orange).}
    \label{fig:pruning}
\end{figure}

Figure~\ref{fig:pruning} illustrates SASP for one memory timestep.

We next detail its computational impact on the SAM2 mask decoder.

\subsection{Setting}
We consider a representative configuration:
\begin{itemize}
    \item $N_o = 10$ tracked objects,
    \item $N_s = 10$ memory timesteps per object (the dual memory bank
    typically ranges from 1 to 12 in practice),
    \item each memory timestep is a $64 \times 64$ grid of tokens
          ($L = 4096$) with embedding dimension $d = 256$,
    \item the query sequence (current-frame image embeddings) has
    $L_q = 4096$ tokens of dimension $d = 256$.
\end{itemize}
For each object, the decoder performs cross-attention from $L_q$ query tokens
to $N_sL$ memory tokens in keys/values $K/V$. Across objects, these operations
are batched.

\paragraph{Baseline cross-attention cost}

Ignoring constant factors (softmax, bias, etc.), the dominant operations in one
cross-attention block are $QK^\top$ and $AV$, where
$Q \in \mathbb{R}^{L_q \times d}$ and
$K,V \in \mathbb{R}^{(N_s L) \times d}$.
The per-object complexity is
\[
\mathcal{O}(L_q N_s L d)
\]
for $QK^\top$; $AV$ has the same order.

For one object with $L_q = 4096$, $L = 4096$, $N_s = 10$, and $d = 256$, the
FLOPs for $QK^\top$ are
\[
\begin{aligned}
\text{FLOPs}_{QK}
&\approx 2 \cdot L_q \cdot (N_s L) \cdot d \\
&\approx 8.6 \times 10^{10}.
\end{aligned}
\]
Since $AV$ has the same order, the total per-object cost is
\[
\text{FLOPs}_{\text{attn,obj}} \approx 1.7 \times 10^{11}.
\]
For $N_o = 10$ objects, one decoder block therefore consumes
\[
\text{FLOPs}_{\text{attn,baseline}} \approx 1.7 \times 10^{12}
\]
FLOPs per frame from cross-attention alone.

Empirically, baseline runtime scales approximately linearly with $N_o$, which
indicates inefficient parallelization for very long attention sequences. After
pruning, batched cross-attention scales sub-linearly with $N_o$, yielding
latency gains beyond raw FLOP reduction.

\subsection{Cost of the Similarity Map}

SASP computes one frame-level cosine-similarity map over encoder embeddings.
For a $64 \times 64$ grid, this is an $L \times L$ matrix with $L = 4096$ and
complexity
\[
\mathcal{O}(L^2 d).
\]
Computing this map once per frame requires
\[
\text{FLOPs}_{\text{sim}} \approx
2 \cdot L^2 \cdot d
\approx 8.6 \times 10^{9}
\]
FLOPs.

Crucially, there is one similarity map per frame. It is computed once, then
reused for all objects and all future steps that still reference that frame in
memory. Previously processed frames already have cached maps and are not
recomputed.

\subsection{Complexity Savings from Pruning}

SASP keeps only (i) tokens on the object mask and
(ii) tokens in sufficiently similar regions of encoder feature space.
Empirically, $\tau_{\text{sim}}$ is set so each timestep retains about $0.4L$
tokens.

The effective number of keys/values per object becomes
\[
L_k' = N_s \cdot 0.4L = 0.4 N_s L,
\]
so the multi-object pruned cross-attention cost is
\[
\text{FLOPs}_{\text{attn, pruned}} \approx 6.8 \times 10^{11}.
\]

Per decoder block and frame, this gives
\begin{align*}
\text{FLOPs saved by pruning}
& \approx 1.0 \times 10^{12}, \\
\text{FLOPs spent on similarity}
&\approx 8.6 \times 10^{9}.
\end{align*}
Thus, similarity-map computation is well below $1\%$ of the FLOPs saved by
attention pruning because it is computed once per frame and reused across
objects.

\begin{table}[t]
\centering
\scriptsize
\setlength{\tabcolsep}{1.1pt}
\renewcommand{\arraystretch}{0.8}
\caption{Measured latencies on DanceTrack, in $ms/frame$.} 
\label{tab:latency_breakdown}
\begin{tabular}{@{}p{0.12\linewidth}p{0.25\linewidth}rrrrrr@{}}
\toprule
Group & Block & \multicolumn{3}{c}{4 obj.} & \multicolumn{3}{c}{30 obj.} \\
\cmidrule(lr){3-5}\cmidrule(l){6-8}
 & & Full & NoDepth & SASP & Full & NoDepth & SASP \\
\midrule
Detector & Detector & 26.5 & 26.5 & 27.2 & 27.5 & 27.9 & 28.0 \\
\midrule
SAM2 & SAM2 & 118.4 & 113.5 & 67.3 & 622.4 & 638.5 & 173.8 \\
\midrule
LiAM & ACM + Bookkeeping & 7.3 & 6.7 & 7.0 & 30.5 & 32.3 & 29.5 \\
LiAM & CTI & 0.1 & 0.1 & 0.1 & 0.5 & 0.5 & 0.5 \\
LiAM & MGG-motion & 3.8 & 3.9 & 4.0 & 11.4 & 11.9 & 10.8 \\
LiAM & MGG-depth & 61.1 & -- & -- & 112.9 & -- & -- \\
LiAM & SASP & -- & -- & 3.4 & -- & -- & 25.8 \\
\midrule
Total & \textbf{SAM2+Detector} & \textbf{144.9} & \textbf{139.9} & \textbf{94.5} & \textbf{650.0} & \textbf{666.5} & \textbf{201.8} \\
Total & \textbf{LiAM} & \textbf{72.2} & \textbf{10.8} & \textbf{14.5} & \textbf{155.2} & \textbf{44.7} & \textbf{66.6} \\
\bottomrule
\end{tabular}
\vspace{-20pt}
\end{table}

\section{Runtime Analysis}
We report measured runtimes of LiAM-SAM on DanceTrack validation sequences using an NVIDIA A100 GPU. Table~\ref{tab:latency_breakdown} provides a per-module latency breakdown across different configurations and numbers of tracked objects. This analysis complements the runtime/accuracy trade-off reported in the main paper by isolating the cost of the detector, SAM2, and the individual LiAM components.
Results show that runtime is dominated by SAM2, especially as the number of tracked objects increases. This confirms that memory-based cross-attention is the primary computational bottleneck. In a crowded scenario with 30 tracked objects, SAM2 component accounts for the largest share of the total latency (77.3\%), motivating the need for  memory-token reduction.
SASP directly targets this bottleneck by pruning spatial memory tokens before cross-attention. As shown in Table~\ref{tab:latency_breakdown}, SASP reduces SAM2 latency by approximately 72\% in the 30-object setting, while introducing only a small additional pruning overhead. This behavior is consistent with the FLOP-level analysis in Sec.~\ref{sec:experiments:pruning}, where reducing the number of retained memory tokens directly lowers the attention cost.
The remaining LiAM components introduce limited overhead relative to SAM2. CTI is negligible, while ACM and bookkeeping scale moderately with the number of active tracks. MGG-motion remains inexpensive and is retained in the fast configuration because it provides the main association benefit. MGG-depth introduces additional cost due to depth estimation and mask refinement, but it is applied selectively and can be disabled when efficiency is prioritized.
Compared with recent transformer-based MOT methods such as ColTrack and DualTemporalMOT, LiAM-SAM operates under a different tracking paradigm that explicitly propagates object masks through SAM2. While this design achieves substantially higher tracking accuracy on both DanceTrack and SportsMOT, as shown in Table~\ref{tab:fps}, it also incurs a lower processing speed and a larger model size than these detector-based MOT approaches. The latency breakdown in Table~\ref{tab:latency_breakdown} shows, however, that most of this overhead originates from SAM2 itself, while the proposed LiAM components contribute only a small fraction of the total runtime.

Overall, the latency breakdown supports the efficiency design of LiAM-SAM: lifecycle-aware memory control improves association robustness with moderate overhead, while SASP mitigates the dominant SAM2 memory-attention cost in crowded scenes.

\begin{table}[ht] \vspace{-8pt}\caption{Comparison steady-state core FPS on DanceTrack validation over 24,940 post-warm-up calls. H/A: HOTA/AssA} \label{tab:fps}
{\centering
\setlength{\tabcolsep}{2.4pt}
\renewcommand{\arraystretch}{0.84}
\fontsize{5.7}{6.2}\selectfont \vspace{-8pt}
\begin{tabular}{lrrrr}
\toprule
Method & DanceTrack H/A & SportsMOT H/A & MParams & FPS on NVIDIA L4 \\
\midrule
ColTrack & 72.6 / 62.3 & 71.5 / 63.6 & 15.8 & \textbf{9.0} \\
Dual-Path & 74.1 / 65.6 & 73.9 / 66.6 & 11.3 & 5.8 \\
LiAM-SAM & \textbf{79.6} / \textbf{72.6} & \textbf{81.8} / \textbf{76.7} & 359.6 & 2.1 \\
\bottomrule
\end{tabular}\par}\vspace{-10pt}
\end{table}

\section{Additional qualitative example}
\begin{figure*}[t]
\vspace{-1em}
\centering

\setlength{\tabcolsep}{2pt}
\renewcommand{\arraystretch}{0.9}

\begin{tabular}{
    @{}C{0.03\textwidth}@{}  
    C{0.02\textwidth}@{}     
    C{0.27\textwidth}@{}      
    C{0.27\textwidth}@{}      
    C{0.27\textwidth}@{}      
}

\multirow{2}{*}{\rotatebox[origin=c]{90}{DanceTrack}}%
  & \rotatebox[origin=c]{90}{Ours}%
  & \includegraphics[width=\linewidth]{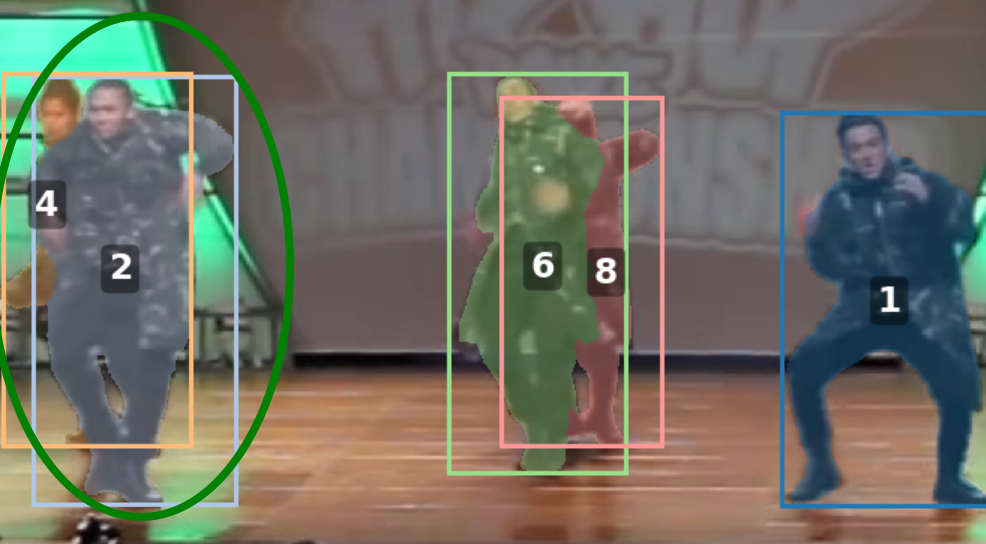}%
  & \includegraphics[width=\linewidth]{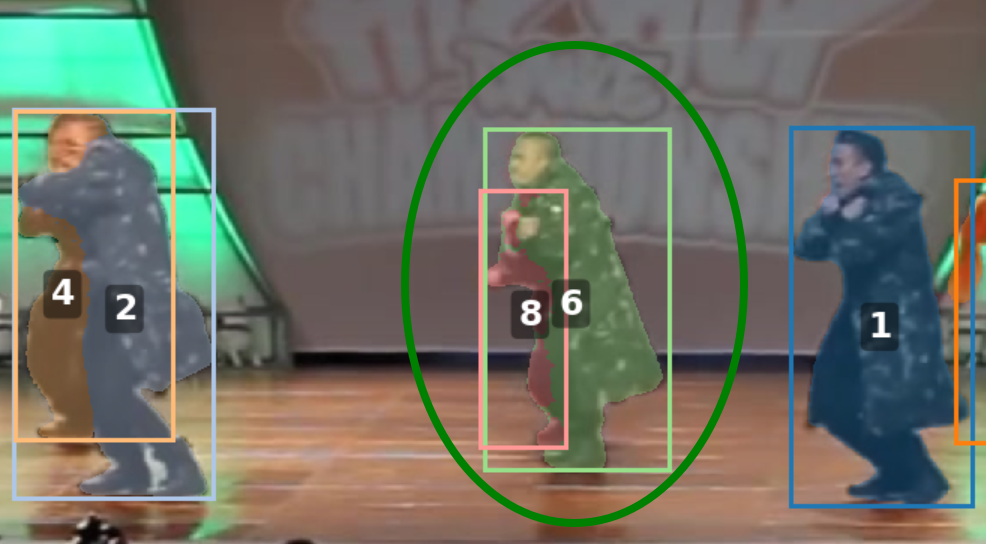}%
  & \includegraphics[width=\linewidth]{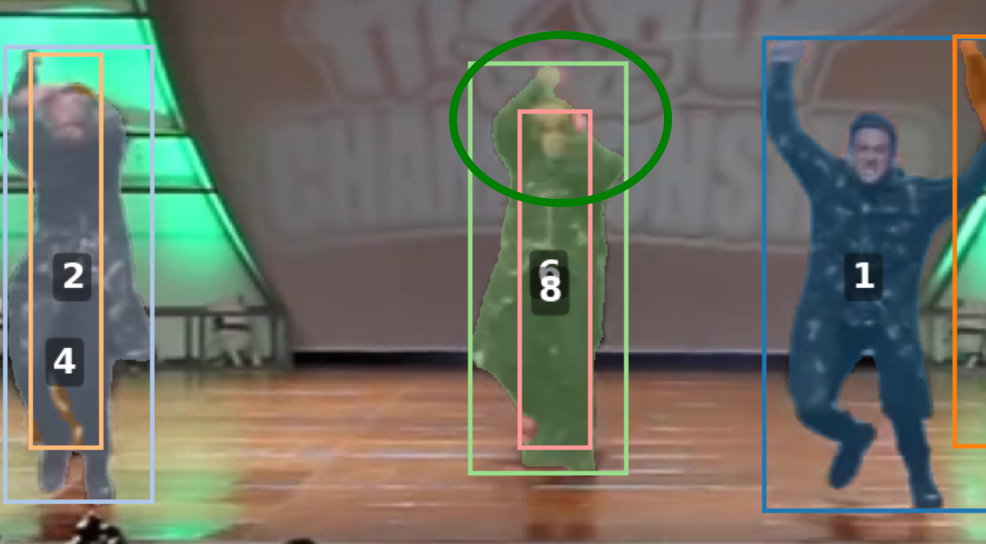}\\[0.0em]

& \rotatebox[origin=c]{90}{Baseline}%
  & \shortstack{%
      \includegraphics[width=\linewidth]{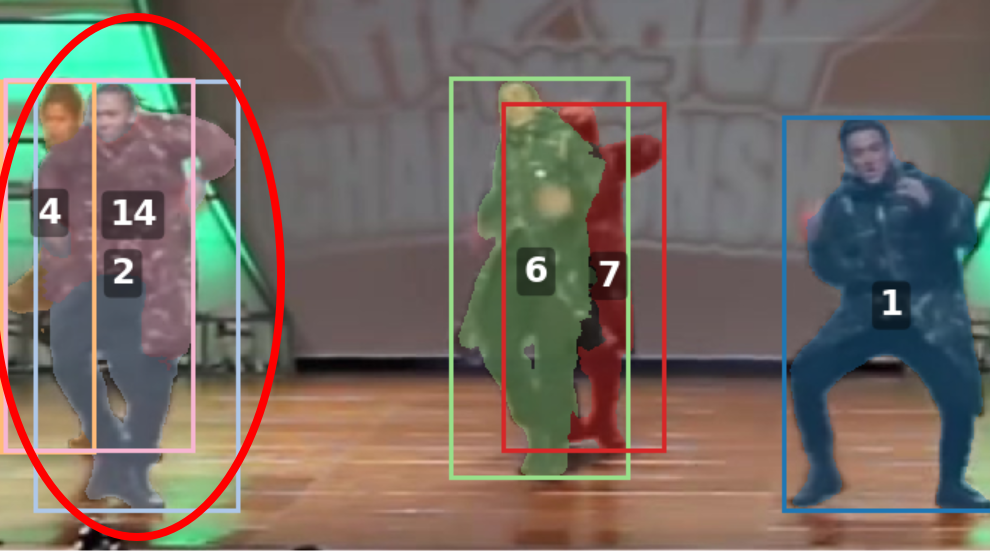}\\[-0.2em]
      {\scriptsize (a) CTI avoids subobjects}%
    }%
  & \shortstack{%
      \includegraphics[width=\linewidth]{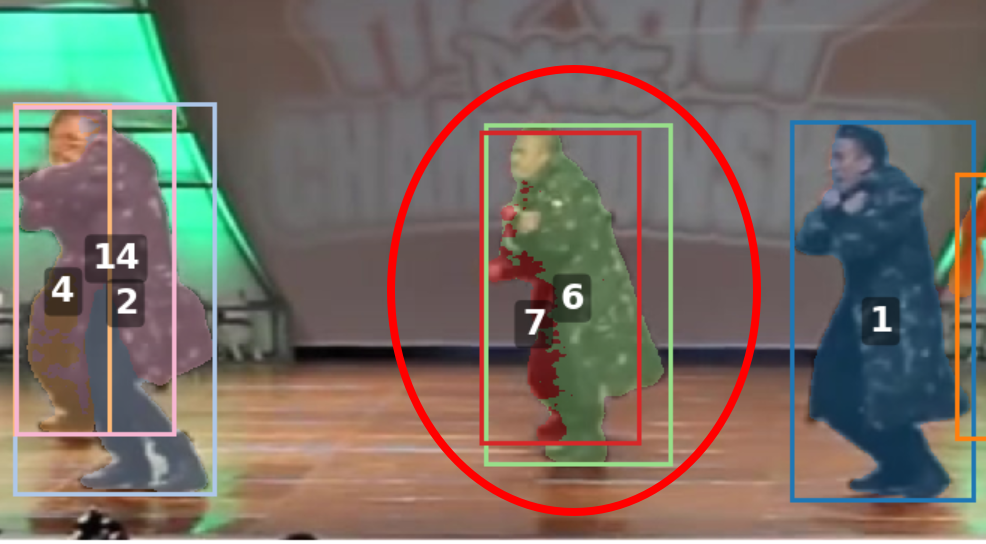}\\[-0.2em]
      {\scriptsize (b) MGG stabilizes masks}%
    }%
  & \shortstack{%
      \includegraphics[width=\linewidth]{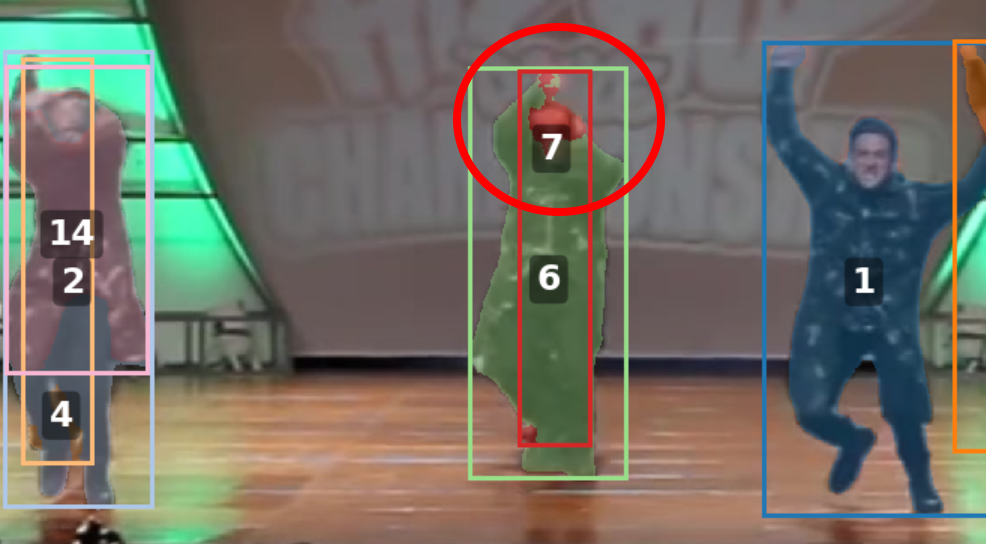}\\[-0.2em]
      {\scriptsize (c) Depth-grounding corrects mask bleeding}%
    }\\
\noalign{\vspace{0.2em}}

\end{tabular}

\caption{Qualitative examples on DanceTrack using MGG-depth}
\label{fig:quality_dance2}
\vspace{-1em}
\end{figure*}

\textit{Complex interactions: cleaner masks and memory.}
Figure \ref{fig:quality_dance2} highlights mask-level failure modes. The baseline initializes spurious ``sub-object'' tracks in (a) and shows unstable boundaries between dancers~\#6 and~\#7 in (b--c). \method prevents sub-object births via CTI and applies depth-guided correction to reduce mask bleeding, yielding cleaner object-specific memory updates.
\nolinenumbers
\linenumbers

\section{Detector fine-tuning details}\label{supp:detector}

For each benchmark, we fine-tune the detector prior to tracking evaluation. Model selection is performed using the Exponential Moving Average (EMA) model, and the checkpoint achieving the highest validation Average Precision (AP) is subsequently used for all experiments on the corresponding dataset. For SportsMOT, detector training includes an additional hard-negative fine-tuning stage. Hard negatives are detections with confidence scores greater than or equal to 0.70 and a maximum Intersection over Union (IoU) below 0.10 with any ground-truth annotation. For all datasets, inference uses a detection confidence threshold of 0.30. Table~\ref{tab:detector_training} summarizes the complete training and model-selection protocols.

\begin{table*}[t] \centering \scriptsize \caption{Detector fine-tuning and model selection protocols.} \label{tab:detector_training} \begin{tabular}{p{0.16\textwidth}p{0.18\textwidth}p{0.43\textwidth}p{0.18\textwidth}} \toprule Dataset & Training Data & Optimization & Model Selection \\ \midrule DanceTrack & Official training split & Resolution 1120; local batch 2 on 8 GPUs (global batch 16); detector/encoder LR $2\times10^{-4}/3.2\times10^{-5}$; weight decay $10^{-4}$; 200 epochs. & EMA COCO AP on the official 25-video validation split selects epoch 1. \\ \midrule BDD100K & Official training split (10\% held out for validation) & Resolution 560; local batch 4 on 8 GPUs (global batch 32); detector/encoder LR $4\times10^{-4}/6.4\times10^{-5}$; weight decay 0.05; 10 epochs. & EMA AP on the hold-out validation set selects epoch 6. \\ \midrule SportsMOT (Stage 1) & Official 45-video training split; one player class & Resolution 1120; local batch 2 on 8 GPUs with accumulation 2 (global batch 32); detector/encoder LR $1.25\times10^{-5}/2\times10^{-6}$; weight decay 0.05; two warm-up epochs; 15 total epochs. & EMA AP selects epoch 10. \\ SportsMOT (Stage 2) & 313 mined hard-negative frames repeated fourfold (29,513 samples) & Detector/encoder LR $2\times10^{-6}/2\times10^{-7}$; weight decay 0.05; 8 epochs. Hard negatives are detections with score $\geq 0.70$ and maximum ground-truth IoU $< 0.10$. & EMA AP selects epoch 3. \\ \bottomrule \end{tabular} \end{table*}

\section{Detailed Implementation}
\label{supp:impl}

This section follows the overall lifecycle update described in
\cref{sec:method} and summarizes the detailed implementation choices 
for each module. Low-level SAM2
memory/decoder settings follow the default pretrained configuration, and the
detector uses the training setup described in \cref{sex:exp:implementation}.
Unless noted otherwise, the same CTI, ACM, and MGG settings are used on
DanceTrack, BDD100K. The main accuracy model does not enable SASP; pruning
is activated only for the runtime ablation.

Algorithm~\ref{alg:liam_update} summarizes the online update using the rules
introduced in \cref{sec:method}, with thresholds instantiated in Table~\ref{tab:sup_impl}.
We write $s(d)$ for the detector score,
$\mathrm{ov}(d,\mathcal{T}_t)$ for the fraction of detection $d$ covered by
already tracked regions, and $\mathrm{dup}(\cdot), \mathrm{sub}(\cdot)$ for
the duplicate and sub-object initialization scores.

\begin{algorithm}[t]
\caption{Per-frame LiAM-SAM update at frame $t$}
\label{alg:liam_update}
\footnotesize
\begin{algorithmic}[1]
\STATE Propagate active tracks with SAM2 and detect boxes with score $\geq \tau_{\mathrm{det}}$
\STATE Associate propagated tracks and detections by Hungarian matching with IoU gate $\tau_{\mathrm{match}}$
\FOR{each unmatched detection $d$}
    \IF{$s(d) < \tau_{\mathrm{birth}}$ or $\mathrm{ov}(d,\mathcal{T}_t) > \tau_{\mathrm{ov}}^{\mathrm{reject}}$}
        \STATE Reject $d$
    \ELSE
        \IF{$\mathrm{ov}(d,\mathcal{T}_t) > \tau_{\mathrm{ov}}^{\mathrm{contrast}}$}
            \STATE Initialize from the detector box with negative contrastive prompts
        \ELSE
            \STATE Initialize from the detector box alone
        \ENDIF
        \IF{$\mathrm{dup}(d) > \tau_{\mathrm{dup}}$}
            \STATE Reject the initialized mask as a duplicate
        \ELSE
            \STATE Add new track
        \ENDIF
    \ENDIF
\ENDFOR
\FOR{each active track $i$}
    \IF{$t \bmod \Delta_{\mathrm{ACM}} = 0$ and $\max_{j \neq i} \mathrm{IoU}(B_i^t,B_j^t) < \tau_{\mathrm{ACM}}$}
        \STATE Promote frame $t$ to ACM and keep at most $K_{\mathrm{ACM}}$ conditioning references
    \ENDIF
\ENDFOR
\FOR{each pair $(i,j)$ with overlapping masks}
    \STATE $q_{ij} \gets \mathrm{IoU}(M_i^t,M_j^t)$
    \IF{$q_{ij} > \tau_{\mathrm{ov}}^{\mathrm{MGG}}$}
        \IF{the score history $H_{\mathrm{score}}$ identifies a suspected loss at cutoff $\tau_{\mathrm{pre}}$}
            \STATE Select that track for suppression
        \ELSE
            \STATE Classify trajectories as close or distinct over $W_{\mathrm{traj}}$ using squared Mahalanobis distance gate $\tau_{d^2}$
            \STATE Route close trajectories to age arbitration and distinct trajectories to score arbitration
            \STATE In score arbitration, use $m_{\mathrm{gap}}$, $m_{\mathrm{drop}}$, and $\tau_{\mathrm{arb}}$; fall back to age if score evidence is inconclusive
        \ENDIF
        \STATE Mark the selected track absent and re-encode its current memory as no-object
        \STATE Retire it after $P_{\mathrm{retire}}$ consecutive confusion suppressions
    \ELSIF{$\tau_{\mathrm{bleed}} < q_{ij} < \tau_{\mathrm{trust}}$}
        \IF{depth coherence $\geq \gamma_{\mathrm{depth}}$ and support separation $\geq \lambda_{\mathrm{sep}}$}
            \STATE Apply depth-guided boundary correction
        \ENDIF
    \ENDIF
\ENDFOR
\IF{SASP is enabled}
    \STATE Dilate each object mask by $r_{\mathrm{dil}}$ and keep tokens with $sim(l,l')$ $\geq \tau_{\mathrm{sim}}$
\ENDIF
\STATE Output masks, boxes, and identities for frame $t$
\end{algorithmic}
\end{algorithm}

\begin{table*}[t]
\centering
\scriptsize
\caption{Main implementation controls for \method. Dataset-specific values are reported as DanceTrack/BDD100K only when they differ; ``off'' denotes a disabled control.}
\label{tab:sup_impl}
\setlength{\tabcolsep}{2pt}
\renewcommand{\arraystretch}{0.95}
\begin{tabular}{@{}>{\raggedright\arraybackslash}p{1.1cm}>{\raggedright\arraybackslash}p{4.75cm}>{\centering\arraybackslash}p{2.3cm}>{\raggedright\arraybackslash}p{8.7cm}@{}}
\toprule
\textbf{Module} & \textbf{Control} & \textbf{DT/BDD} & \textbf{Operational meaning} \\
\midrule
Assoc. & Detection-score floor $\tau_{\mathrm{det}}$ & $0.30$ & Minimum detector score retained before association. \\
Assoc. & Matching-IoU gate $\tau_{\mathrm{match}}$ & $0.30$ & Minimum detection-to-track box IoU accepted during association. \\
\midrule
CTI & Birth-score threshold $\tau_{\mathrm{birth}}$ & $0.60/0.50$ & Minimum detector score accepted for a new-track candidate. \\
CTI & Birth-rejection tracked coverage $\tau_{\mathrm{ov}}^{\mathrm{reject}}$ & $0.55/0.90$ & Candidate-box fraction covered by active tracks above which birth is rejected. \\
CTI & Contrastive-prompt tracked coverage $\tau_{\mathrm{ov}}^{\mathrm{contrast}}$ & $0.35$ & Candidate-box fraction covered by active tracks above which negative prompts are added. \\
CTI & Post-init duplicate-mask overlap $\tau_{\mathrm{dup}}$ & $0.03$ & Initialized-mask fraction covered by active tracks above which the new track is rejected. \\
\midrule
ACM & Reference-update interval $\Delta_{\mathrm{ACM}}$ & $10$ & Frame interval between candidate reference promotions. \\
ACM & Reference capacity $K_{\mathrm{ACM}}$ & $6$ & Maximum number of conditioning references exposed to attention. \\
ACM & Reference-promotion overlap gate $\tau_{\mathrm{ACM}}$ & $0.50$ & Maximum pairwise box IoU allowed for reference promotion. \\
\midrule
MGG-Mo & Interaction-overlap gate $\tau_{\mathrm{ov}}^{\mathrm{MGG}}$ & $0.80$ & Predicted-mask IoU above which motion-grounded overlap resolution runs. \\
MGG-Mo & Score-history window $H_{\mathrm{score}}$ & $8$ & Number of past object-score frames used by suspected-loss and score arbitration. \\
MGG-Mo & Score-gap / score-drop margins $(m_{\mathrm{gap}},m_{\mathrm{drop}})$ & $4$ & Minimum current-score gap and historical score drop used for score arbitration. \\
MGG-Mo & Suspected-loss score cutoff $(\tau_{\mathrm{pre}},\tau_{\mathrm{arb}})$ & $2$ & Current-score cutoff before trajectory routing and within score arbitration. \\
MGG-Mo & Trajectory-routing window $W_{\mathrm{traj}}$ & $8$ & Recent trajectory window used to route between age and score arbitration. \\
MGG-Mo & Squared Mahalanobis distance gate $\tau_{d^2}$ & $6$ & Maximum trajectory distance $d^2$ for treating two trajectories as close. \\
MGG-Mo & Confusion-retirement patience $P_{\mathrm{retire}}$ & $2$ & Consecutive confusion suppressions before a track is deactivated. \\
\midrule
MGG-De & Bleed-correction overlap floor $\tau_{\mathrm{bleed}}$ & $0.05$ & Minimum overlap required to activate depth-guided bleed correction. \\
MGG-De & Depth-trust overlap ceiling $\tau_{\mathrm{trust}}$ & $0.45$ & Maximum overlap for which depth-guided boundary correction is trusted. \\
MGG-De & Depth-coherence threshold $\gamma_{\mathrm{depth}}$ & $0.60$ & Minimum depth coherence required for boundary correction. \\
MGG-De & Depth-support separation $\lambda_{\mathrm{sep}}$ & $0.50/2.00$ & Minimum separation between the two depth-support distributions. \\
\midrule
SASP & Mask-dilation radius $r_{\mathrm{dil}}$ & $1$ & Object-mask dilation applied before memory-token selection. \\
SASP & Token-similarity threshold $\tau_{\mathrm{sim}}$ & $0.45$ & Minimum cosine similarity for retaining a memory token. \\
\bottomrule
\end{tabular}
\end{table*}

\end{document}